\pdfoutput=1
\documentclass[runningheads]{llncs}
\usepackage{graphicx}
\usepackage{comment}
\usepackage{amsmath,amssymb} % define this before the line numbering.
\usepackage{color}
\usepackage{xcolor}
\usepackage{microtype}
\usepackage{scalerel}
\usepackage{multirow}
\usepackage{booktabs}
\usepackage[caption=false]{subfig}
\usepackage{xfrac}
\usepackage{algorithm}
\usepackage{algorithmic}
\usepackage{enumitem}

\begin{document}

\pagestyle{headings}
\mainmatter
\def\ECCVSubNumber{5696}  % Insert your submission number here

\title{Exploiting Temporal Coherence for Self-Supervised One-shot Video Re-identification} % Replace with your title

\begin{comment}
\titlerunning{ECCV-20 submission ID \ECCVSubNumber} 
\authorrunning{ECCV-20 submission ID \ECCVSubNumber} 
\author{Anonymous ECCV submission}
\institute{Paper ID \ECCVSubNumber}
\end{comment}
%******************

%\begin{comment}
\titlerunning{Exploiting Temporal Coherence for Self-Supervised One-shot Video Re-ID}
\author{Dripta S. Raychaudhuri \and
Amit K. Roy-Chowdhury}
\authorrunning{D. S. Raychaudhuri and A. K. Roy-Chowdhury}
\institute{University of California, Riverside\\
\email{\{draychaudhuri,amitrc\}@ece.ucr.edu}}
%\end{comment}
%******************
\maketitle

%%% ABSTRACT
\begin{abstract}
While supervised techniques in re-identification are extremely effective, the need for large amounts of annotations makes them impractical for large camera networks. One-shot re-identification, which uses a singular labeled tracklet for each identity along with a pool of unlabeled tracklets, is a potential candidate towards reducing this labeling effort. Current one-shot re-identification methods function by modeling the inter-relationships amongst the labeled and the unlabeled data, but fail to fully exploit such relationships that exist within the pool of unlabeled data itself. In this paper, we propose a new framework named Temporal Consistency Progressive Learning, which uses temporal coherence as a novel self-supervised auxiliary task in the one-shot learning paradigm to capture such relationships amongst the unlabeled tracklets. Optimizing two new losses, which enforce consistency on a local and global scale, our framework can learn learn richer and more discriminative representations. Extensive experiments on two challenging video re-identification datasets - MARS and DukeMTMC-VideoReID - demonstrate that our proposed method is able to estimate the true labels of the unlabeled data more accurately by up to $8\%$, and obtain significantly better re-identification performance compared to the existing state-of-the-art techniques.

\end{abstract}

%%% INTRODUCTION
\section{Introduction}
Person re-identification (re-ID) aims to solve the challenging problem of matching identities across non-overlapping views in a multi-camera system. The surge of deep neural networks in computer vision \cite{krizhevsky2012imagenet,ren2015faster} has been reflected in person re-ID as well, with impressive performance over a wide variety of datasets \cite{guangcong2019aaai,Chen_2019_ICCV}. However, this improved performance has predominantly been achieved through \textit{supervised learning}, facilitated by the availability of large amounts of annotated data. However, acquiring identity labels for a large set of unlabeled tracklets is an extremely time-consuming and cumbersome task. Consequently, methods which can ameliorate this annotation problem and work with limited supervision, such as \textit{unsupervised learning} or \textit{semi-supervised learning} techniques, are of primary importance in the context of person re-ID. 

\begin{figure}[t]
   \centerline{
    \includegraphics[width=\textwidth]{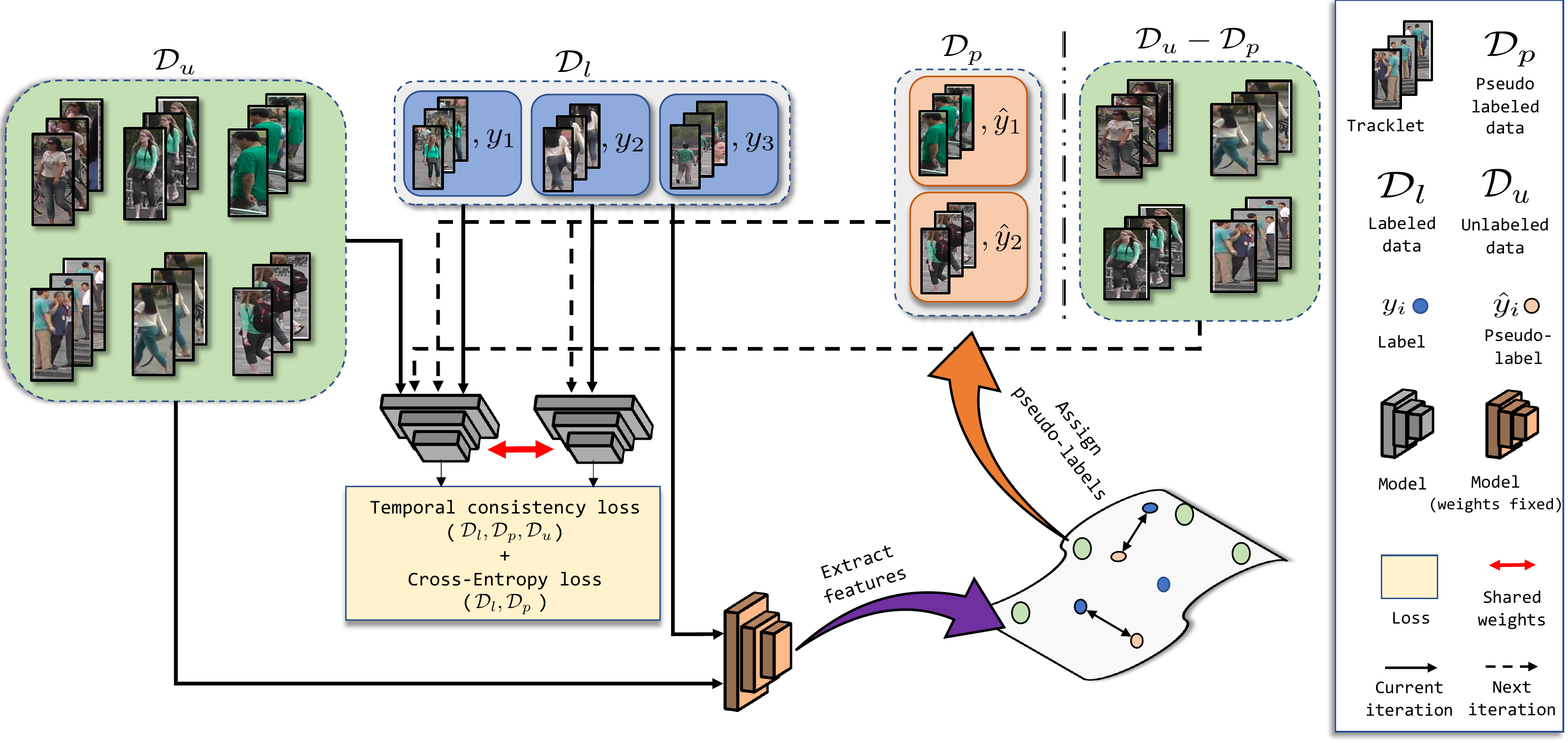}}
    \caption{\textbf{A schematic illustration of the proposed framework}. Our method makes use of both labeled and unlabeled tracklets at every iteration of model training. The first step involves learning the parameters of the deep model by using temporal consistency as self-supervision and, additionally, softmax loss on the minimal set of annotated tracklets. Next, this model is used to predict pseudo-labels on a few confident samples. These two steps alternate, one after the other, until the entire unlabeled set has been incorporated in terms of pseudo-labels.}
    \label{fig:concept}
\end{figure}
In this work, we focus on the semi-supervised task in video person re-ID, specifically, the \textit{one-shot} setting, where only one tracklet per identity is labeled. The objective of the learning process is to utilize this small labeled set along with a larger unlabeled set of tracklets to obtain a re-ID model. The key challenge involved with the one-shot task is figuring out the inter-relationships which exist amongst the labeled and unlabeled instances. State-of-the-art one-shot methods try to address this by estimating the labels of the unlabeled tracklets (pseudo-labels) and then utilizing a supervised learning strategy. Some works employ a static sampling strategy \cite{ye2017dynamic,liu2017stepwise}, where pseudo-labels with a confidence score above a pre-defined threshold are selected for supervised learning. More recent works \cite{wu2018cvpr_oneshot,wu2019progressive} make use of a progressive sampling strategy, where a subset of the pseudo-labeled samples are selected with the size of the subset expanding with each iteration. This prevents an influx of noisy pseudo-labels, and thus, averts the situation of confirmation bias \cite{arazo2019pseudolabeling}. However, in an effort to control the number of noisy pseudo-labels, most of these methods discard a significant portion of the unlabeled set at each learning iteration; thus, the information in the unlabeled set is not maximally utilized for training the model. Due to this inefficient usage of the unlabeled set and the limited number of labeled instances, propagating beliefs directly from the labeled to the unlabeled set is insufficient to fully capture the relationships which exist amongst instances of the unlabeled set.

To resolve this issue of inefficient usage of the unlabeled data, we draw inspiration from the field of self-supervised visual representation learning \cite{kolesnikov2019revisiting}. We propose using \emph{temporal coherence} \cite{paul2018incorporating,mobahi2009deep,misra2016shuffle} as a form of self-supervision to maximally utilize the unlabeled data and learn discriminative person specific representations. Temporal coherence is motivated by the fact that features corresponding to a person in a tracklet should be focused on the discriminative aspects related to the person, such as clothing and gait, and ignore background nuances such as illumination and occlusion (see Fig. \ref{fig:inter}). This naturally suggests that features should be temporally consistent across the entire duration of the tracklet as the person in a tracklet remains constant. Thus, we propose a new framework, \emph{Temporally Consistent Progressive Learning} (TCPL), which unifies this notion of temporal coherence with a progressive pseudo-labeling strategy \cite{wu2018cvpr_oneshot}. An overview of our framework is presented in Fig. \ref{fig:concept}.

In this paper, we propose two novel losses to learn such temporally consistent features: \emph{Intra-sequence temporal consistency loss} and the \emph{Inter-sequence temporal consistency loss}. Both of these losses apply consistency regularization on the temporal dimension of a tracklet. While the first loss employs a local level of consistency by operating on a specific tracklet, the second loss extends it by applying  temporal consistency both \emph{within and across} tracklets.

Using such self-supervised losses, our framework can use the unlabeled data at each iteration of learning, allowing maximal information to be extracted out of it. Additionally, by exploiting two levels of consistency, as explained above, TCPL can better model the relationships amongst the unlabeled instances without being limited by the number of labeled instances. Thus, our framework addresses both the drawbacks associated with the current crop of methods and achieves state-of-the-art performance in the one-shot person re-ID task.
\paragraph{\bf{Main contributions.}}
Our main contributions are summarised as follows: 
\begin{itemize}
    \item[$\bullet$] We introduce a new framework, \emph{Temporally Consistent Progressive Learning}, which unifies self-supervision and pseudo-labeling to maximally utilize the labeled and unlabeled data efficiently for one-shot video person re-ID.
    \item[$\bullet$] We introduce two novel self-supervised losses, the \emph{Intra-sequence temporal consistency loss} and the \emph{Inter-sequence temporal consistency loss}, to implement temporal consistency and empirically demonstrate their benefits in learning richer and more discriminative feature representations. 
    \item [$\bullet$] We demonstrate that this intelligent use of the unlabeled data through self-supervision, unlike previous pseudo-labeling methods, leads to significantly better label estimation and superior results on the one-shot video re-ID task, outperforming the state-of-the art one-shot video re-ID methods on the MARS and DukeMTMC-VideoReID datasets.
\end{itemize}

%%% RELATED WORKS
\section{Related works}
The majority of the literature in person re-ID has focused on \emph{supervised} learning on labeled images/tracklets of persons \cite{zhou2019osnet,zheng2016person,chen2018video,xu2017jointly}. While these techniques achieve excellent results on many datasets, they require a substantial amount of annotations. The need to alleviate this excessive need for labeled data has led to research into \emph{unsupervised} \cite{yu2017cross,yu2019unsupervised,lin2019bottom,chen2018bmvc} and \emph{semi-supervised} \cite{wu2018cvpr_oneshot,wu2019progressive,ding2019feature} methods. We provide a review of the relevant developments in these fields. In addition, our work draws inspiration from the ideas explored in the domain of \emph{self-supervision}. \\
\noindent \textbf{Unsupervised person re-ID.} Recent unsupervised methods \cite{yu2017cross,yu2019unsupervised,lin2019bottom,chen2018bmvc} mostly use some form of deep clustering. The authors in \cite{li2019unsupervised} utilise a camera aware loss by defining nearest neighbors across cameras as being similar. In \cite{lin2019bottom}, an agglomerative clustering scheme is introduced, alternating between learning of features and clustering using the learnt features. However, these methods still lag behind supervised methods by quite some distance. Another line of research utilises auxiliary datasets, which are completely labeled, for initializing a re-ID model and then using unsupervised domain adaptation techniques on the unsupervised target dataset. \\
\noindent \textbf{Semi-supervised \& one-shot person re-ID.} The unsatisfactory performance of purely unsupervised methods \cite{yu2017cross,yu2019unsupervised,lin2019bottom,chen2018bmvc} has given rise to semi-supervised and one-shot methods in re-ID. Some of the major ideas utilized in these methods include dictionary learning \cite{liu2014semi}, graph matching \cite{hamid2016joint} and metric learning \cite{bak2017one}. More recently, new methods in this setting try to estimate the labels of the unlabeled tracklets (pseudo-labels) with respect to the labeled tracklets and then utilise a supervised learning strategy. The authors of \cite{ye2017dynamic} use a dynamic graph matching strategy which iteratively updates the image graph and the label estimation to learn a better feature space with intermediate estimated labels. A stepwise metric learning approach to the problem is proposed in \cite{liu2017stepwise}. Both these methods employ a static sampling strategy, where pseudo-labels with a confidence score above a pre-defined threshold are selected at each step - this leads to a lot of noisy labels being incorporated and hinders the learning process due to due to \textit{confirmation bias} \cite{arazo2019pseudolabeling}. In order to contain the noise, the authors of \cite{wu2018cvpr_oneshot,wu2019progressive} approach the problem from a progressive pseudo-label selection strategy, where the subset of the pseudo-labeled samples selected gradually increase with iterations. While this prevented the influx of noisy pseudo-labels, a significant portion of the unlabeled set is discarded at each step and thus, the unlabeled set is used inefficiently. We address this issue by using self-supervision.\\
\noindent \textbf{Self-supervised learning.} Self-supervised learning utilizes pretext tasks, formulated using only unsupervised data. A pretext task is designed in a such a way that solving it requires the model to learn useful visual features. These tasks can involve predicting the angle of rotation applied to an image \cite{gidaris2018unsupervised} or predicting a permutation of multiple randomly sampled and permuted patches \cite{noroozi2016unsupervised}. Some techniques go beyond solving such auxiliary classification tasks and enforce constraints on the representation space. A prominent example is the exemplar loss from \cite{dosovitskiy2014discriminative}. Our method belongs to this latter category of self-supervision and imposes temporal consistency on tracklet features.

%%% METHODOLOGY
\section{Methodology}
In this section, we present our framework (TCPL) for solving the task of one-shot video person re-ID. First, we provide a background on the current progressive pseudo-labeling methods and discuss their shortcomings. Thereafter, we turn to our proposed temporal consistency losses and describe their workings, before presenting our integrated framework. Before going into the details of our framework, let us define the notations and problem statement formally.
\paragraph{\bf{Problem statement.}} Consider that we have a training set of $m$ tracklets, $\mathcal{D}=\{\mathcal{X}_i\}_{i=1}^m$, which are acquired from a camera network. One-shot re-ID assumes that there exists a set $\mathcal{D}_l \subset \mathcal{D}$, which contains a singular labeled tracklet for each identity. Thus, $\mathcal{D}_l=\{(\mathcal{X}_i,y_i)\}_{i=1}^{m_l}$, where $y_i \in \{0,1\}^{m_l}$ such that $y_i$ is $1$ only at dimension $i$ and $0$ otherwise, and $m_l$ denotes the number of distinct identities. The rest of the tracklets, $\mathcal{D}_u=\mathcal{D}-\mathcal{D}_l=\{\mathcal{X}_i\}_{i=1}^{m_u}$ do not possess annotations. Our goal is to learn a discriminative person re-ID model $f_{\theta}(\cdot)$ utilizing both $\mathcal{D}_l$ and $\mathcal{D}_u$. During inference, $f_{\theta}(\cdot)$ is used to embed both the probe $\mathcal{X}^\mathsf{q}$ and gallery tracklets $\{\mathcal{X}_i^\mathsf{g}\}_{i=1}^{m_g}$ into a common space and then rank all the gallery tracklets by evaluating their degree of correspondence to the probe via some metric. What makes this challenging, even more so than the semi-supervised task, is the fact that $m_l \ll m_u$ and each identity has only a single labeled tracklet. 

\subsection{Progressive Pseudo-labeling and its drawbacks} \label{Progressive Learning}
The progressive pseudo-labeling paradigm is an enhancement over the original pseudo-labeling framework \cite{lee2013pseudo} where one imputes approximate
classes on unlabeled data by making predictions from a model trained only on labeled data. The learning process involves the following two steps for each step of learning: (1) train the model via supervised learning on the labeled data and the pseudo-labeled data; (2) select a few reliable pseudo-labeled candidates from unlabeled data according to a prediction reliability criterion.

In \cite{wu2018cvpr_oneshot}, the authors gradually select larger sets of pseudo-labeled data to be incorporated into the supervised learning process via a dissimilarity criterion. Pseudo labels are assigned to the unlabeled candidates by the identity labels of their nearest labeled neighbors in the embedding space. The distance to the corresponding labeled neighbor is designated as the dissimilarity cost, which is used as the measure of reliability for the pseudo label. However, as a result of the strict selection criterion, this does not use the unlabeled set efficiently - discarding a significant amount of unlabeled data at each step of pseudo labeling.

To improve the efficiency, the authors in \cite{wu2019progressive} propose to set up a memory bank to store the instance features $v_i = f_{\theta}(\mathcal{X}_i)$ calculated in the previous step. Then the probability of sample $\mathcal{X}_j$ being recognized as the $i$-th instance can be written as,
\begin{equation} \label{softmax}
    P(i|\mathcal{X}_j) = \frac{\exp{(v_i^Tf_{\theta}(\mathcal{X}_j)/\tau)}}{\sum_k\exp{(v_k^Tf_{\theta}(\mathcal{X}_j)/\tau)}}
\end{equation}
where $\tau$ is the temperature parameter controlling the softness of the distribution. Minimizing the negative log-likelihood of $\sum_i P(i|\mathcal{X}_j)$, which they call the \textit{exclusive loss}, pulls each instance $\mathcal{X}_i$ towards its corresponding memorized vector $v_i$ and repels the memorized vectors of other instances. Due to efficiency issues, the memorized feature $v_i$ corresponding to instance $\mathcal{X}_i$ is only updated in the iteration which takes $\mathcal{X}_i$ as input \cite{ye2019unsupervised}. In other words, the memorized feature $v_i$ is only updated once per epoch. However, the network itself is updated in each iteration, rendering the memory bank scheme inefficient. In addition, the exclusive loss looks at the global data distribution, similar to the softmax loss, forcing embeddings corresponding to different identities to stay apart for encouraging inter-class separability. The local data distribution or the intra-class similarity, is left unaddressed and thus, the improvement over softmax is negligible. 

In the next section, we present how temporal coherence can be employed to amend these drawbacks.

\subsection{Temporal coherence as self-supervision}
In the previous section, we discussed the two fundamental problems plaguing the current crop of progressive pseudo-labeling methods: (1) inefficient usage of the unlabeled set, (2) focusing strictly on the global data distribution. To ameliorate these drawbacks, \emph{we propose to use temporal coherence as a form of self-supervision}. Consistency across the frames in a tracklet encourages the model to focus on the \emph{local} distribution of the data and learn features which incorporate the specific attributes of the individual in the tracklet and ignore spurious artifacts such as background and lighting variation. This also provides a straightforward approach towards utilizing the entire unlabeled set, irrespective of whether some specific unlabeled instance is assigned a confident pseudo-label. In the following sections, we present two novel losses: \emph{Intra-sequence temporal consistency} and \emph{Inter-sequence temporal consistency}, which implement this notion of temporal consistency and show how to integrate them into a self-learning framework towards solving the one-shot video re-ID task.

\subsubsection*{Intra-sequence temporal consistency.}
The intra-sequence temporal consistency loss is based on the idea of video temporal coherence \cite{paul2018incorporating,mobahi2009deep,misra2016shuffle}. While the previous works focus on learning the temporal order by considering individual frames, we use consistency as a tool for the learnt features to implicitly \textit{ignore background nuances} and \textit{focus on the actual person attributes}. We do this by sampling non-overlapping mini-tracklets from a tracklet and enforce the embeddings corresponding to these mini-tracklets to come closer via a contrastive loss. 

Given a tracklet $\mathcal{X}$ consisting of frames $\{x_1,\cdots,x_n\}$, intra-sequence consistency involves creating two mini-tracklets $\mathcal{X}^{\mathsf{a}}$ and $\mathcal{X}^{\mathsf{p}}$ by sampling two mutually exclusive sets of frames from the original tracklet $\mathcal{X}$. This is done by the function $\mathrm{\Phi}_\mathsf{T}(\mathcal{X})$, which first divides the $\mathcal{X}$ into a set of mini-tracklets, each of size $\rho\cdot|\mathcal{X}|$ and then samples from it as follows,
\begin{align}
    \mathcal{X}^\mathsf{a}, \mathcal{X}^\mathsf{p} = \mathrm{\Phi}_{\mathsf{T}}(\mathcal{X})
\end{align}
More specifically, $\mathrm{\Phi}_\mathsf{T}(\mathcal{X})$ samples from the set $\{\mathcal{X}^1,\mathcal{X}^2,\cdots,\mathcal{X}^{1/\rho}\}$ uniformly without replacement. Here, $\rho$ is a hyper-parameter that controls the size of each mini-tracklet with respect to the size of the tracklet $|\mathcal{X}|$. This ensures that $\mathcal{X}^{\mathsf{a}}\cap\mathcal{X}^{\mathsf{p}}=\emptyset$, and consequently, these tracklets are temporally incoherent. For all our experiments, $\rho$ is set to $0.2$. After obtaining these tracklets the loss forces their respective representations to be consistent temporally with one another as follows,
\begin{equation} \label{isc}
    \mathcal{L}_{\text{intra}} = \Vert f_{\theta}(\mathcal{X}^\mathsf{a}) - f_{\theta}(\mathcal{X}^\mathsf{p})) \Vert_2.
\end{equation}
This definition of the intra-sequence temporal consistency can be interpreted as a from of consistency regularization \cite{miyato2018virtual,tarvainen2017mean,oliver2018realistic}, which measures discrepancy between predictions made on perturbed unlabeled data points, i.e.,
\begin{equation} \label{const}
    \mathcal{L}_{\text{cons}} = d\left(p(y|x),p(y|\hat{x})\right)
\end{equation}
where $d(\cdot,\cdot)$ is a divergence measure and $\hat{x}=x+\delta$. Such regularization focuses on the local data distribution, and implicitly pushes the decision boundary away from high-density parts of the unlabeled data to enhance intra-class similarity in accordance to the \textit{cluster assumption} \cite{chapelle2009semi}. In our formulation, the two mini-tracklets are \emph{temporally perturbed versions of each other in terms of background}, i.e., $x=\mathcal{X}^\mathsf{a}$, $\hat{x}=\mathcal{X}^\mathsf{p}$ and $\delta$ indicates perturbations in time - the consistency is applied on features, instead of distributions, and across time. 

\subsubsection*{Inter-sequence temporal consistency.}
The intra-sequence temporal consistency loss focuses solely on the intra-class similarity. To learn a discriminative person re-ID model, the learning process also has to account for the global distribution of the data or the inter-class separability. The triplet loss \cite{hermans2017defense} has been widely used in the re-identification and retrieval literature for its ability to encode such global information. 

The triplet loss ensures that, given an anchor point $\mathcal{X}^\mathsf{a}$, the feature of a positive point $\mathcal{X}^\mathsf{p}$ belonging to the same class (person) $y_a$ is closer to the feature of the anchor than that of a negative point $\mathcal{X}^\mathsf{n}$ belonging to another class $y_n$, by at least a margin $\alpha$. However, directly using the triplet loss is not possible in our scenario as it uses identity label information and thus, its effectiveness will depend heavily on the quality of label estimation. Therefore, we propose the inter-sequence temporal consistency loss, which induces a global level of consistency similar to the standard triplet formulation \emph{without access to labels}. 

Specifically, given a tracklet $\mathcal{X}$, we sample two temporally incoherent mini-tracklets in the same manner as mentioned in the previous section. Without loss of generality, we treat one as the anchor $\mathcal{X}^\mathsf{a}$, and the other one as  the positive point $\mathcal{X}^\mathsf{p}$, which contains the same identity, but temporally perturbed. For the negative instance, we obtain it from the batch nearest neighbors of $\mathcal{X}^\mathsf{a}$. This is done by creating the corresponding ranking list of tracklets in the batch B, excluding $\mathcal{X}$ and sampling a tracklet $\mathcal{X}^\mathsf{n}$ uniformly within the range of ranks $[r,2r]$ as follows: 
\begin{align}
    \mathcal{X}^\mathsf{n} = \mathrm{\Psi}(\mathrm{\mathcal{N}}_{[r,2r]}(\mathcal{X}))
\end{align}
where $\mathrm{\Psi}(\cdot)$ denotes sampling from a set of elements uniformly. $\mathcal{N}_{[r,2r]}(\mathcal{X})$ indicates the nearest neighbors of $\mathcal{X}$ in the batch (up to a total of B neighbors) which are ranked in the range $[r,2r]$. Using this range of ranks we filter out the possible positive samples and the easy negative samples, which are very low in the ranking list and potentially contribute to zero gradient. This strategy allows us to choose potential hard negatives which have been shown to give best performance \cite{hermans2017defense}. The value of $r$ is set to $3$ and $\alpha$ to $0.3$, for all our experiments.

\begin{figure}[t]
    \centering
    \includegraphics[width=\textwidth]{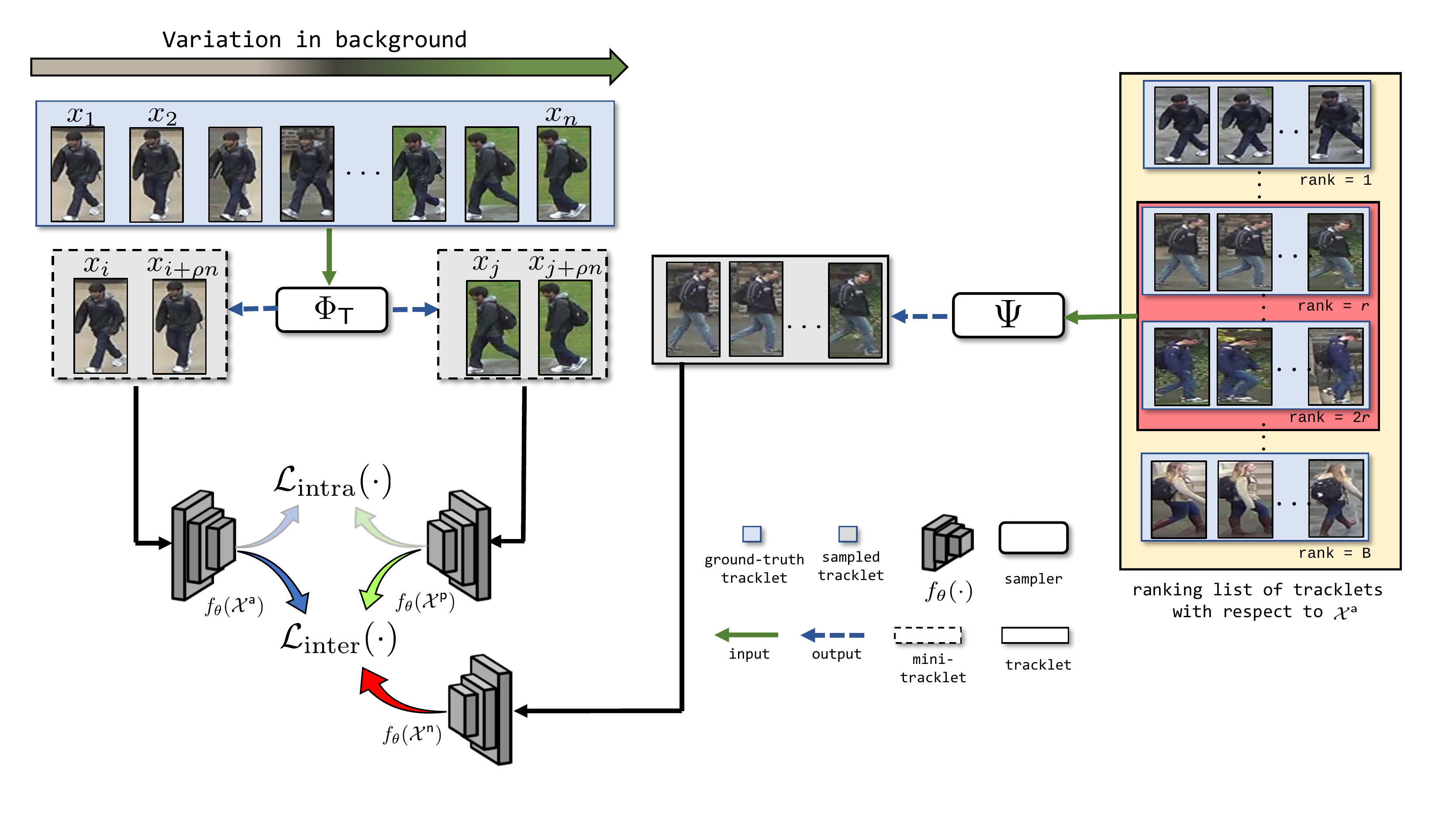}
    \caption{\textbf{An illustration of the inter-sequence temporal consistency criterion}. Firstly, we sample temporally incoherent mini-tracklets using $\mathrm{\Phi_{\mathsf{T}}}$ to serve as the anchor and positive sample. Note the temporal perturbations in these mini-tracklets, manifested in the form changing background. Next, $\mathrm{\Psi}$ is used to obtain the negative sample from the batch nearest neighbors of the anchor, using a ranking based criterion. Using these, we formulate the triplet loss to enforce consistency such that $f_{\theta}(\cdot)$ learns features which focus on the discriminative aspects related to the person in the tracklet and ignore the background nuances.}
    \label{fig:inter}
\end{figure}

Thus, the inter-sequence temporal consistency loss can be formulated as, 
\begin{equation} \label{nnt}
     \mathcal{L}_{\text{inter}} = \text{max}\left\{0, \Vert f_{\theta}(\mathcal{X}^\mathsf{a}) - f_{\theta}(\mathcal{X}^\mathsf{p})\Vert_2 - \Vert f_{\theta}(\mathcal{X}^\mathsf{a}) - f_{\theta}(\mathcal{X}^\mathsf{n}) \Vert_2 + \alpha\right\}
\end{equation}
A pictorial representation of the loss formulation is presented in Fig. \ref{fig:inter}.

\subsection{Temporal Consistency Progressive Learning}
In this section, we present our proposed framework, \emph{Temporal Consistency Progressive Learning} (TCPL), based on the temporal consistency self-supervised losses discussed in the previous section. TCPL integrates self-supervision with pseudo-labeling to learn the person re-ID model. Temporal coherence is used to enhance the feature learning process in the form of multi-task learning. Training of this framework alternates between two key steps: (1) Representation learning, (2) Assignment of pseudo-labels. 

\paragraph{\bf{Representation learning.}} In order to learn the weights of the embedding function $f_{\theta}(\cdot)$, we jointly optimize the following loss function,
\vspace{-1.5mm}
\begin{multline} \label{loss}
    \mathcal{L}=\sum_{\scaleto{(\mathcal{X},y) \in \mathcal{D}_l}{5pt}}\mathcal{L}_{l}(\mathcal{X},y) + \sum_{\scaleto{(\mathcal{X},\hat{y}) \in \mathcal{D}_p}{5pt}}\mathcal{L}_{l}(\mathcal{X},\hat{y})
    + \lambda\left(\sum_{\scaleto{\mathcal{X} \in \mathcal{D}}{3.5pt}}\mathcal{L}_{\text{intra}}(\mathcal{X})+\sum_{\scaleto{\mathcal{X} \in \mathcal{D}}{3.5pt}}\mathcal{L}_{\text{inter}}(\mathcal{X})\right)
\end{multline}
where $\mathcal{L}_l$ is a standard cross-entropy classification loss applied on
all labeled and selected pseudo-labeled tracklets in the dataset. The supervised loss $\mathcal{L}_l$ is optimized by appending a classifier $g_W(\cdot)$ on top of the feature extractor $f_{\theta}(\cdot)$ as 
\begin{align}
    \mathcal{Z} = g_W(&f_{\theta}(\mathcal{X})) = W^Tf_{\theta}(\mathcal{X})+b \\
    \mathcal{L}_l &= -\text{log}\left(\frac{e^{y^T\mathcal{Z}}}{\sum_je^{\mathcal{Z}_j}}\right),
\end{align}
where $f_{\theta}(\mathcal{X})\in \mathbb{R}^{d\times1}$, $W\in\mathbb{R}^{d\times m_l}$ and $b \in \mathbb{R}^{m_l\times1}$. The value of $d$ represents the feature dimension and is equal to 2048 in our experiments. The labeled set and pseudo-labeled set are denoted by $\mathcal{D}_l$ and $\mathcal{D}_p$ respectively, with $\hat{y}$ denoting the pseudo-labels, while $\mathcal{D}$ refers to the entire set of tracklets. Note that, $\mathcal{D}_l \subset \mathcal{D}$ and $\mathcal{D}_p \subset \mathcal{D}$, such that $\mathcal{D}_p \cap \mathcal{D}_l = \emptyset$. The hyper-parameter $\lambda$ is a non-negative scalar that controls the weight of temporal consistency in the joint loss function.

\paragraph{\bf{Assignment of pseudo-labels.}} Following \cite{wu2018cvpr_oneshot}, we use the nearest neighbor in the embedding space to assign pseudo-labels -  each unlabeled tracklet is assigned a pseudo-label by transferring the label of its nearest labeled neighbor in the embedding space. For $\mathcal{X}_j \in \mathcal{D}_u$,
\begin{align} \label{pseudo}
i = \arg\min_{\mathcal{X}_k \in \mathcal{D}_l} &\Vert f_{\theta}(\mathcal{X}_j) - f_{\theta}(\mathcal{X}_k) \Vert_2, \\
&\hat{y}_j = y_i
\end{align}
After assignment of the pseudo-labels, a confidence criterion is used to choose the most reliable predictions to be used in optimizing $\mathcal{L}_l$ for the next step. Instead of a static threshold, a total of $n_t$ samples are selected at step $t$ by choosing the top $n_t$ unlabeled samples with smallest distance to their corresponding labeled nearest neighbour and added to $\mathcal{D}_p$. A smaller value of the distance implies a more confident pseudo-label prediction. 

The value of $n_t$ is incremented gradually with $t$, depending on an enlarging factor $p \in (0,1)$ \cite{wu2018cvpr_oneshot} where, $n_t = n_{t-1} + pn_u$. Thus, the learning process continues for a total of $\left( \left\lfloor{1/p}\right\rfloor +1\right)$ steps - until the entire unlabeled set has been assigned confident pseudo-labels. The parameter $p$ controls the trade-off between label estimation accuracy and training time - a smaller value of $p$ leads to better label estimation at the cost of higher training time.

\begin{algorithm}
\caption{Temporally Consistent Progressive Learning}

\begin{flushleft}
        \textbf{INPUT:} Labeled set $\mathcal{D}_l$, unlabeled set $\mathcal{D}_u$, enlarging ratio $p$, sampling factor $\rho$, loss weight $\lambda$, randomly initialized model $f_{\theta_0}(\cdot)$\\
        \textbf{OUTPUT:} Feature extractor $f_{\theta_{opt}}(\cdot)$
\end{flushleft}
\begin{algorithmic}[1]
\STATE Initialize the selected pseudo-labeled data $\mathcal{D}_p^0\leftarrow\emptyset$, step $t \leftarrow 0$, sampling size $n_0 \leftarrow 0$, $n_u = |\mathcal{D}_u|$
\WHILE{$n_t \leq n_u$}
\STATE $t \leftarrow t + 1$
\STATE Train the model using (\ref{loss})
\STATE Assign pseudo-labels using (\ref{pseudo})
\STATE $n_t \leftarrow n_{t-1} + p.n_u$
\STATE Choose the $n_t$ most confident pseudo-labels and add to $\mathcal{D}_p^{t-1}$
\ENDWHILE
\STATE Choose model with best validation performance 
\end{algorithmic}
\label{algo}
\end{algorithm}

%%% EXPERIMENTS
\section{Experiments}
We evaluate our proposed method on two popular video person re-ID benchmarks, namely, MARS \cite{zheng2016mars} and DukeMTMC-VideoReID \cite{ristani2016performance}. MARS is the largest video re-ID dataset containing $17,503$ tracklets for $1,261$ identities and $3,248$ distractor tracklets, which are captured by six cameras. The
DukeMTMC-VideoReID dataset is captured using 8 cameras and contains $2,196$ tracklets for training and $2,636$ tracklets for testing. Standard splits are used along with distractors.

{{\it Evaluation metrics.}} Given a probe tracklet, we calculate the Euclidean distance with respect to all the gallery tracklets, and sort the distances to obtain the final ranking list. We utilize the Cumulative Matching Characteristics (CMC) and mean Average Precision (mAP) as the performance evaluation measures. We
report the Rank-1, Rank-5, Rank-20 scores to represent the CMC curve.

{{\it Initial data selection.}} To initialize the labeled and unlabeled sets, we follow the protocol outlined in \cite{wu2018cvpr_oneshot}. For each identity, a tracklet is chosen randomly in camera 1. If camera 1 does not record an identity, a tracklet in the next available camera is chosen to ensure each identity has one tracklet for initialization.

{{\it Implementation details.}} Please see supplementary material for details on implementation, values of different hyper-parameters and datasets.
% \paragraph{{\bf Implementation details}} We use PyTorch \cite{paszke2017automatic} for all experiments. For our model, we use a ResNet-50 \cite{he2016deep} pre-trained on ImageNet \cite{deng2009imagenet} - the last classification layer removed and a fully-connected layer with batch normalization \cite{ioffe2015batch} and a classification layer are added at the end of the model. We adopt stochastic gradient descent (SGD)  with momentum $0.5$ and weight decay $0.0005$ to optimize the parameters for $70$ epochs, with batch size $16$ in each iteration. We set $\lambda=1$ in Eqn. \ref{loss} for the DukeMTMC-VideoReID dataset and $\lambda=0.8$ for the MARS dataset (due to the huge disparity in the number of labeled and unlabeled tracklets as a result of fragmentation in MARS). The learning rate is initialized to $0.1$. In the last 15 epochs, to stabilize the model training and prevent overfitting, we change the learning rate to $0.01$ and set $\lambda = 0$. We provide our implementation in the supplementary material.

\begin{table*}[t]
\setlength{\tabcolsep}{7pt}
\caption{Comparison of TCPL with state-of-the-art one-shot and unsupervised methods on the MARS and DukeMTMC-VideoReID datasets. (Sup./Unsup. refers to supervised and unsupervised methods respectively.)}
\begin{center}
\begin{tabular}{@{}llllllll@{}}
\toprule[1.2pt]
\multirow{2}{*}{Method} & \multirow{2}{*}{Setting} & \multicolumn{3}{c}{MARS}  & \multicolumn{3}{c}{Duke}  \\ \cmidrule(l){3-8} 
                        &                          & R-1  & R-5   & mAP  & R-1  & R-5   & mAP  \\ \midrule[1.2pt]
Baseline: upper bound         & Sup. & 80.8 &92.1 & 67.4 & 83.6 & 94.6 & 78.3 \\ \midrule
TCPL -full (Ours)                   & 1-shot                   & \bf{65.2} & \bf{77.5}  & \bf{43.6} & \bf{76.8} & \bf{87.8}  & \bf{67.9} \\
TCPL -$\mathcal{L}_{\text{intra}}$ (Ours) & 1-shot & 63.3 & 75.2 & 42.9 & 76.2 & 87.6 & 67.7\\
TCPL -$\mathcal{L}_{\text{inter}}$ (Ours) & 1-shot & 64.9 & 77.5 & 43.1 & 74.4 & 86.6 & 66.5\\
\midrule
One-Shot Prog. \cite{wu2019progressive}         & 1-shot                   & 62.8 & 75.2  & 42.6 & 72.9 & 84.3  & 63.3 \\
EUG  \cite{wu2018cvpr_oneshot}                   & 1-shot                   & 62.7 & 72.9  & 42.5 & 72.8 & 84.2  & 63.2 \\
Stepwise Metric  \cite{liu2017stepwise}       & 1-shot                   & 41.2 & 55.6  & 19.7 & 56.3 & 70.4  & 46.8 \\ 
DGM+IDE  \cite{ye2017dynamic}               & 1-shot                   & 36.8 & 54.0  & 16.9 & 42.4 & 57.9  & 33.6 \\
Baseline: lower bound & 1-shot & 36.2 & 50.2 & 15.5 & 39.6 & 56.8 & 33.3\\ \midrule
BUC \cite{lin2019bottom}                    & Unsup.                   & 61.1 & 75.1     & 38.0 & 69.2 & 81.1   & 61.9 \\
UTAL \cite{li2019unsupervised}                   & Unsup.                   & 49.9 & 66.4  & 35.2 & -    & -        & -    \\
DAL  \cite{chen2018bmvc}                   & Unsup.                  & 46.8 & 63.9  & 21.4 & -    & -    & -      \\ \bottomrule
\end{tabular}
\end{center}
\label{table:perf}
\end{table*}

\paragraph{{\bf Comparison to the State-Of-The-Art methods}.}
One-shot re-ID methods in the literature can be broadly divided into two classes: (1) DGM \cite{ye2017dynamic} and Stepwise Metric \cite{liu2017stepwise} use the entire pseudo-labeled data at each step of learning and in the process incorporate a lot of noisy labels, (2) EUG \cite{wu2018cvpr_oneshot} and One-Example Progressive Learning \cite{wu2019progressive} employ progressive sampling. TCPL outperforms all of these by learning an embedding which is temporally consistent. We also consider two baselines: Baseline (one-shot), which utilizes only the one-shot data for training, and Baseline(supervised), which assumes all the tracklets in the training set are labeled; these are trained in a supervised manner using only the cross-entropy loss. We also compare against state-of-the-art unsupervised methods which report results on video re-ID datasets: BUC \cite{lin2019bottom}, UTAL \cite{li2019unsupervised} and DAL \cite{chen2018bmvc}.

We present the results for different instantiations of our framework in Table \ref{table:perf}: one which uses both the losses (TCPL -full) and two others corresponding to usage of the losses individually (TCPL -$\mathcal{L}_{\text{intra}}$,TCPL -$\mathcal{L}_{\text{inter}}$). For TCPL, EUG \cite{wu2018cvpr_oneshot} and One-Shot Progressive \cite{wu2019progressive}, we set the enlarging parameter $p$ to $0.05$.The consistency losses lead to consistent gains of in both rank-$1$ accuracy and mAP over both EUG \cite{wu2018cvpr_oneshot} and One-Shot Progressive Learning \cite{wu2019progressive} in both the datasets. 

\begin{figure*}[b]
\centering
\subfloat[]{
	\label{subfig:mars1}
	\includegraphics[width=0.23\textwidth]{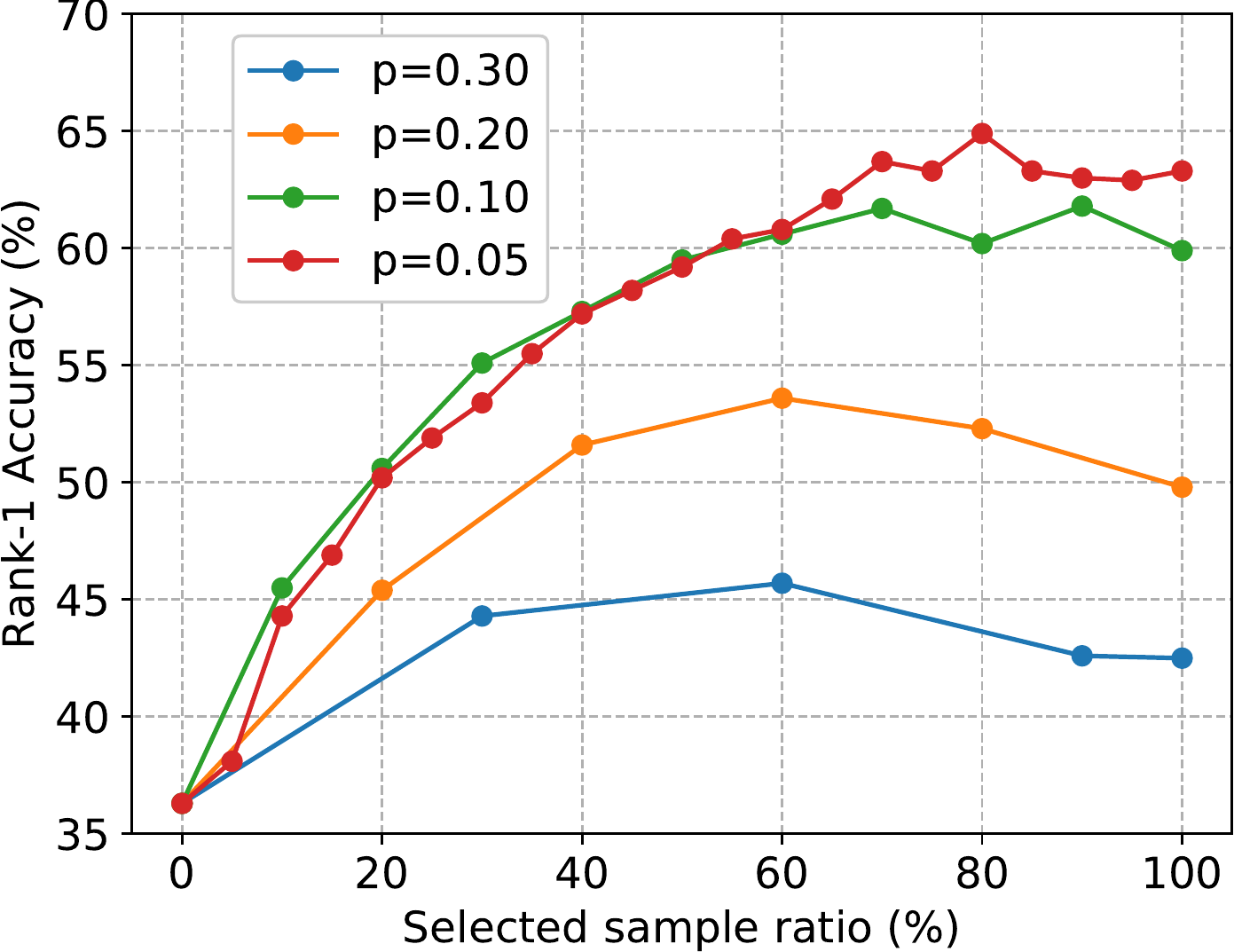} } 
\hfill
\subfloat[]{
	\label{subfig:mars2}
	\includegraphics[width=0.23\textwidth]{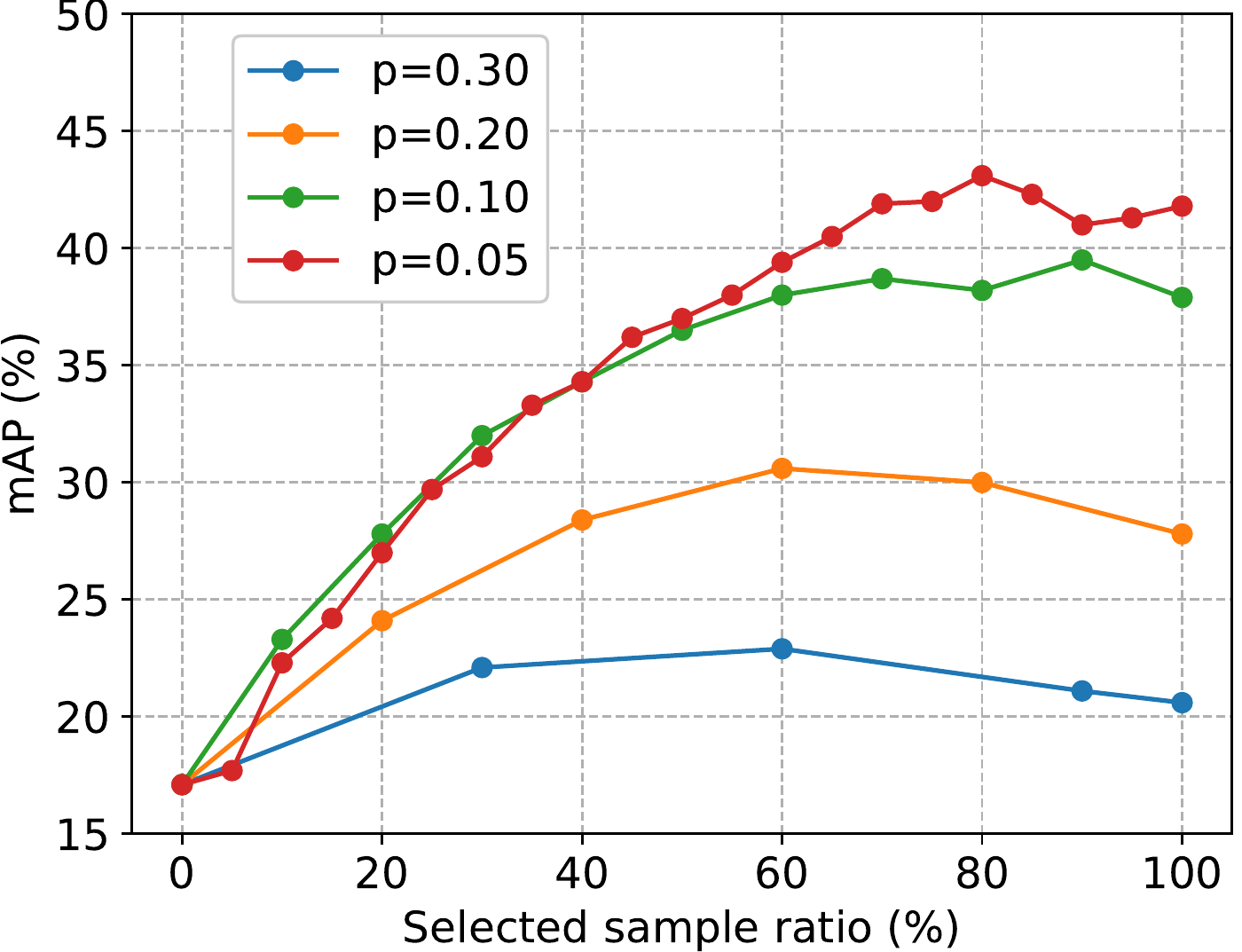} } 
\hfill
\subfloat[]{
	\label{subfig:mars3}
	\includegraphics[width=0.23\textwidth]{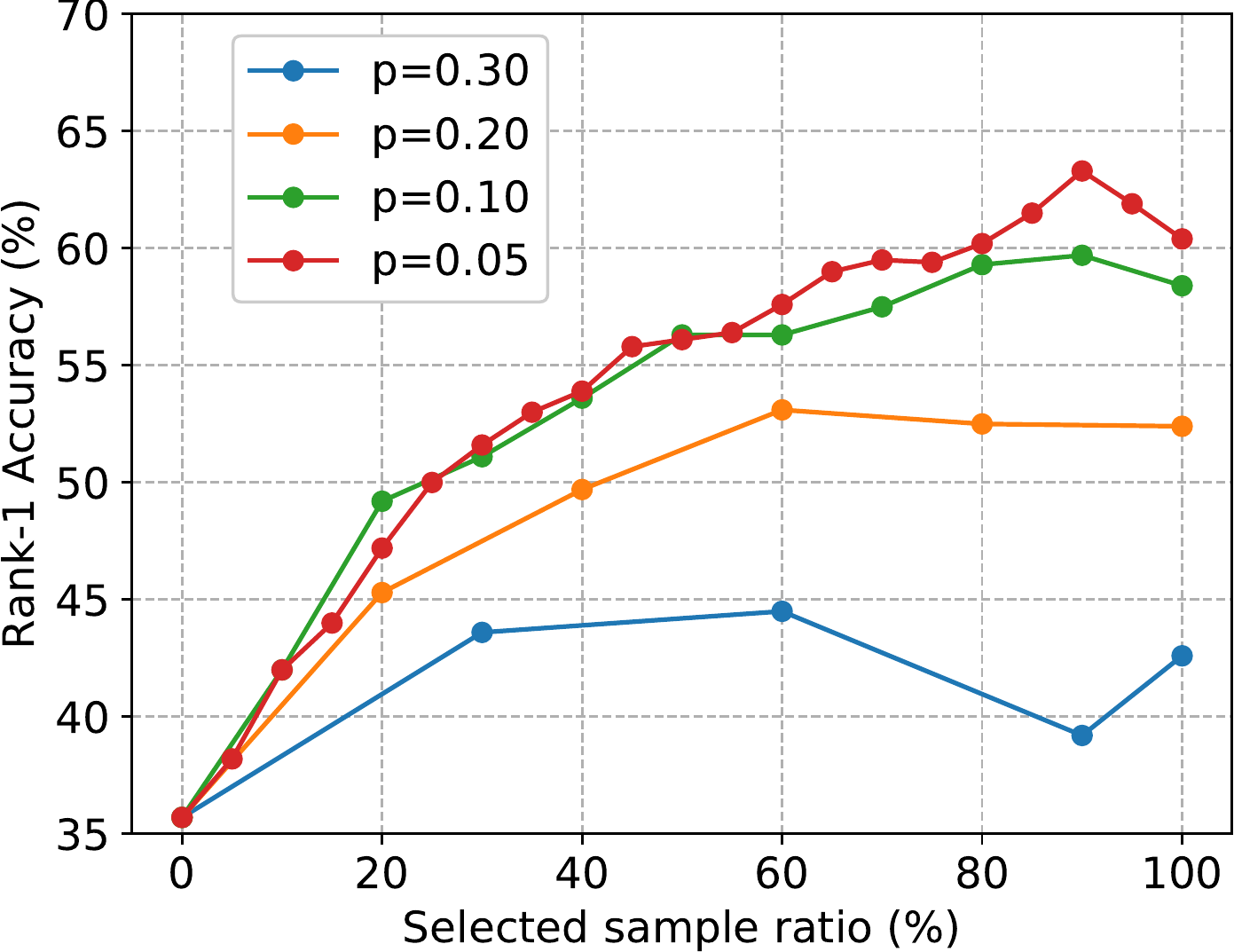} } 
\hfill
\subfloat[]{
	\label{subfig:mars4}
	\includegraphics[width=0.23\textwidth]{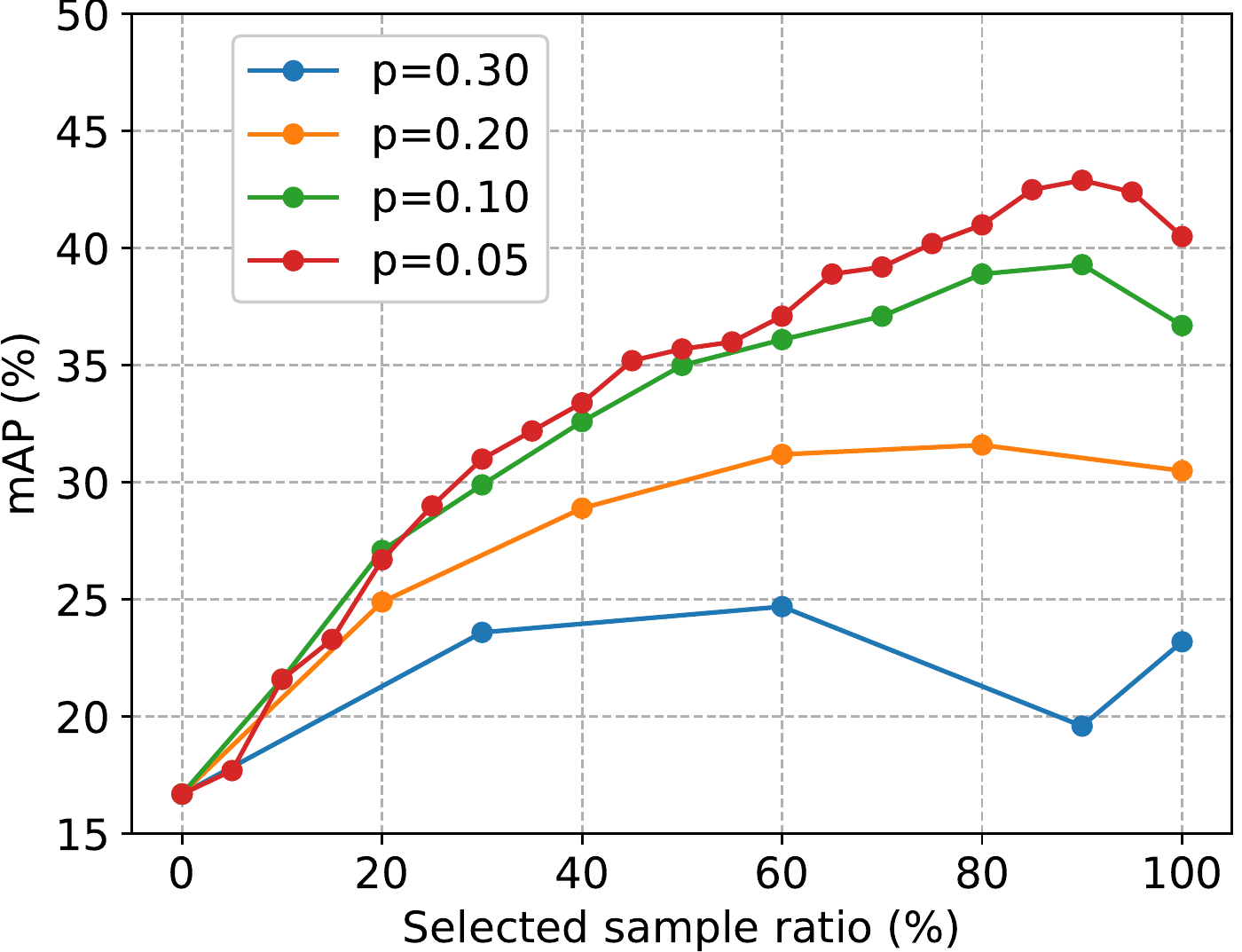} } 
\caption{{\bf Comparison with different values of enlarging factor on MARS.} Figures (a) and (b) represent the Rank-1 accuracy and mAP using TCPL with $\mathcal{L}_{\text{inter}}$. Figures (c) and (d) represent the Rank-1 accuracy and mAP using TCPL with $\mathcal{L}_{\text{intra}}$.}
\label{fig:learning_curves}
\end{figure*}

\paragraph{{\bf Analysis over enlarging factor $p$}.} The selection of the enlarging factor $p$ plays an important role in progressive sampling methods. Decreasing the value of $p$ generally leads to less label estimation errors due to careful data selection, at the cost of a very slow learning process (See Fig. \ref{fig:learning_curves}).

The performance of our method as $p$ varies is shown in Table \ref{table:p_results}. Unlike baseline methods, which suffer drastic drops in performance as $p$ is increased, our framework limits label estimation errors via the consitency losses. Notably, TCPL at $p=0.20$ is able to outperform both EUG and One-Shot Progressive Learning at $p=0.05$ on DukeMTMC-VideoReID. This translates to a $\mathbf{4\times}$ speedup of learning without sacrificing performance. On MARS, at $p=0.10$, TCPL is able to achieve a Rank-1 accuracy of $61.8\%$. This is only $1\%$ behind One-Shot Progressive Learning with $p=0.05$ and suggests a $\mathbf{2\times}$ speedup with only a negligible drop in performance. All of these indicate that TCPL is robust to appending pseudo-labeled data more aggressively and thus, can save time. 

\setlength{\tabcolsep}{5pt}
\begin{table}[t]

\caption{Variation in one-shot performance results for different scales of the enlarging parameter $p$. The best and second best results are in \textcolor{red}{red}/\textcolor{blue}{blue} respectively.}

\begin{center}
\begin{tabular}{lcllllllll@{}}
\toprule[1.2pt]
\multirow{2}{*}{$p$} & \multirow{2}{*}{Method} & \multicolumn{4}{c}{Duke}  & \multicolumn{4}{c}{MARS}  \\ \cmidrule(l){3-10}       &       & R-1  & R-5  & R-20 & mAP   & R-1 & R-5 & R-20 & mAP\\ 
\midrule[1.2pt]
\addlinespace
\multirow{4}{*}{$0.20$} & EUG  \cite{wu2018cvpr_oneshot}                 & 68.9 & 81.1 & 89.4 & 59.5 & 48.7 & 63.4 & 72.6 & 26.6\\
                         & One-Shot Prog. \cite{wu2019progressive} & 69.1 & 81.2 & \textcolor{blue}{89.6} & 59.6 & 49.6 & 64.5 & \textcolor{blue}{74.4} & 27.2\\
                         & TCPL -$\mathcal{L}_{\text{intra}}$           & \textcolor{red}{74.4} & \textcolor{red}{85.8} & \textcolor{red}{91.6} & \textcolor{red}{65.4} & \textcolor{blue}{52.5} & \textcolor{blue}{65.6} & 
                          73.9 & 
                         \textcolor{red}{31.6} \\
                         & TCPL -$\mathcal{L}_{\text{inter}}$          & \textcolor{blue}{69.4} & \textcolor{blue}{81.6} & 
                         88.5 & 
                         \textcolor{blue}{60.5} & \textcolor{red}{53.6} & \textcolor{red}{66.2} & \textcolor{red}{74.9} & \textcolor{blue}{30.6}\\ \addlinespace
                         \midrule[0.6pt]\addlinespace
\multirow{4}{*}{$0.10$} & EUG \cite{wu2018cvpr_oneshot}                 & 70.8 & 83.6 & 89.6 & 61.8 & 57.6 & 69.6 & 78.1 & 34.7\\
                         & One-Shot Prog. \cite{wu2019progressive} & 71.0 & 83.8 & 90.3 & 61.9 & 57.9 & 70.3 & 79.3 & 34.9\\
                         & TCPL -$\mathcal{L}_{\text{intra}}$            & \textcolor{blue}{74.8} & \textcolor{red}{87.3} & \textcolor{blue}{92.0} & \textcolor{blue}{66.7} &
                         \textcolor{blue}{59.7} & \textcolor{blue}{72.0} & \textcolor{blue}{79.3} & \textcolor{blue}{39.3}\\
                         & TCPL -$\mathcal{L}_{\text{inter}}$            & \textcolor{red}{74.9} & \textcolor{blue}{86.5} & \textcolor{red}{92.0} & \textcolor{red}{67.2} &
                         \textcolor{red}{61.8} & \textcolor{red}{74.7} & \textcolor{red}{81.5} & \textcolor{red}{39.5}\\ \addlinespace
                         \midrule[0.6pt]\addlinespace
\multirow{4}{*}{$0.05$} & EUG \cite{wu2018cvpr_oneshot}                 & 72.8 & 84.2 & 91.5 & 63.2 & 62.7 & 72.9 & 82.6 & 42.5\\
                         & One-Shot Prog. \cite{wu2019progressive}& 72.9 & 84.3 & 91.4 & 63.3 & 62.8 & 75.2 & \textcolor{blue}{83.8} & 42.6\\
& TCPL -$\mathcal{L}_{\text{intra}}$            & \textcolor{red}{76.2} & \textcolor{red}{87.6} & \textcolor{red}{92.9} & \textcolor{red}{67.7} & \textcolor{blue}{63.3} & \textcolor{blue}{75.2} & 82.4 & \textcolor{blue}{42.9} \\
                         & TCPL -$\mathcal{L}_{\text{inter}}$           & \textcolor{blue}{74.4} & \textcolor{blue}{86.6} & \textcolor{blue}{92.2} & \textcolor{blue}{66.5} &
                         \textcolor{red}{64.9} & \textcolor{red}{77.5} & \textcolor{red}{84.1} & \textcolor{red}{43.1}\\\bottomrule
\end{tabular}
\end{center}
\label{table:p_results}
\end{table}

\begin{figure}[ht]
\centering
\subfloat[]{
	\label{subfig:ablation_map}
	\includegraphics[width=0.48\textwidth]{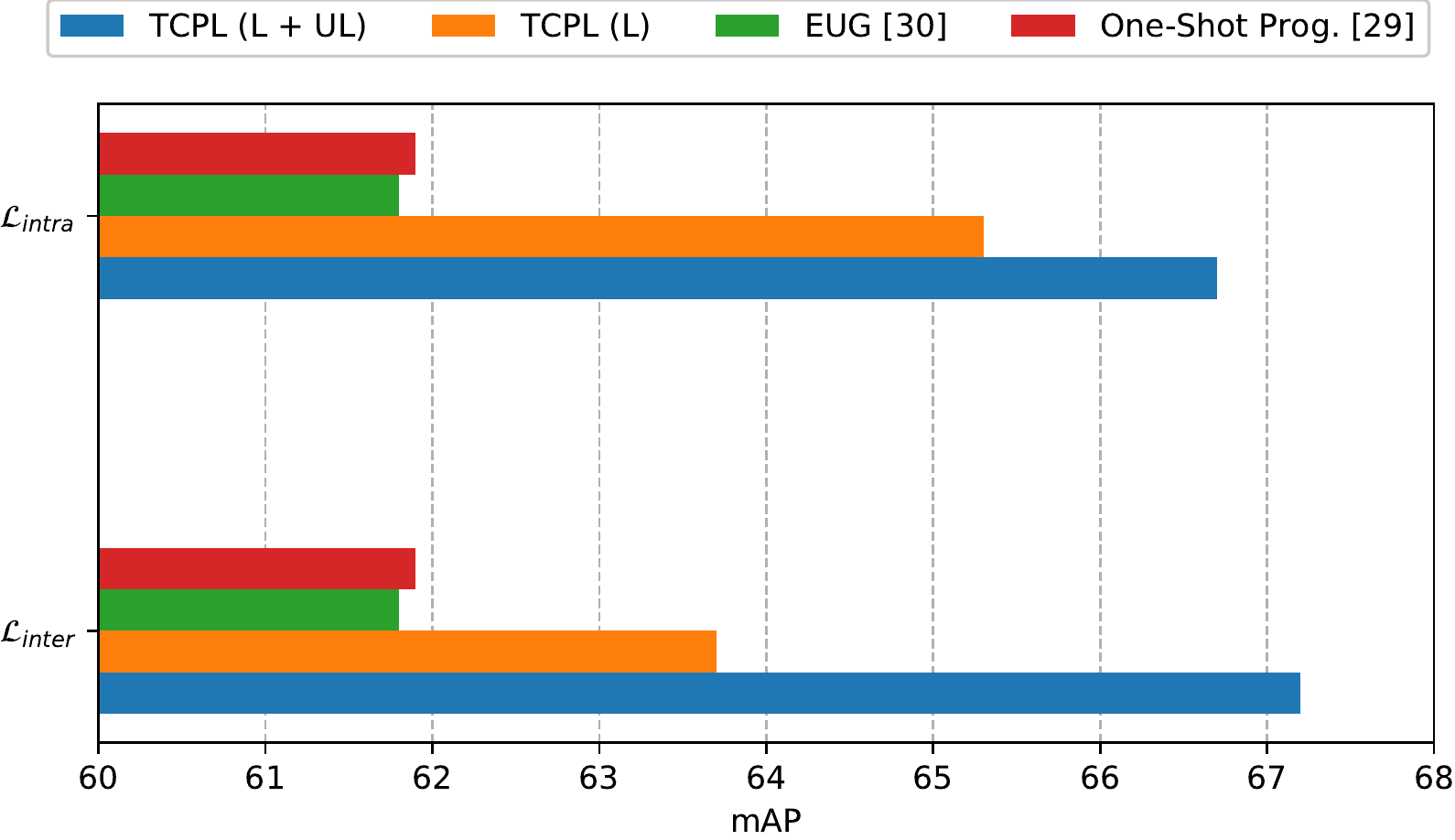} } 
\hfill
\subfloat[]{
	\label{subfig:ablation_r1}
	\includegraphics[width=0.48\textwidth]{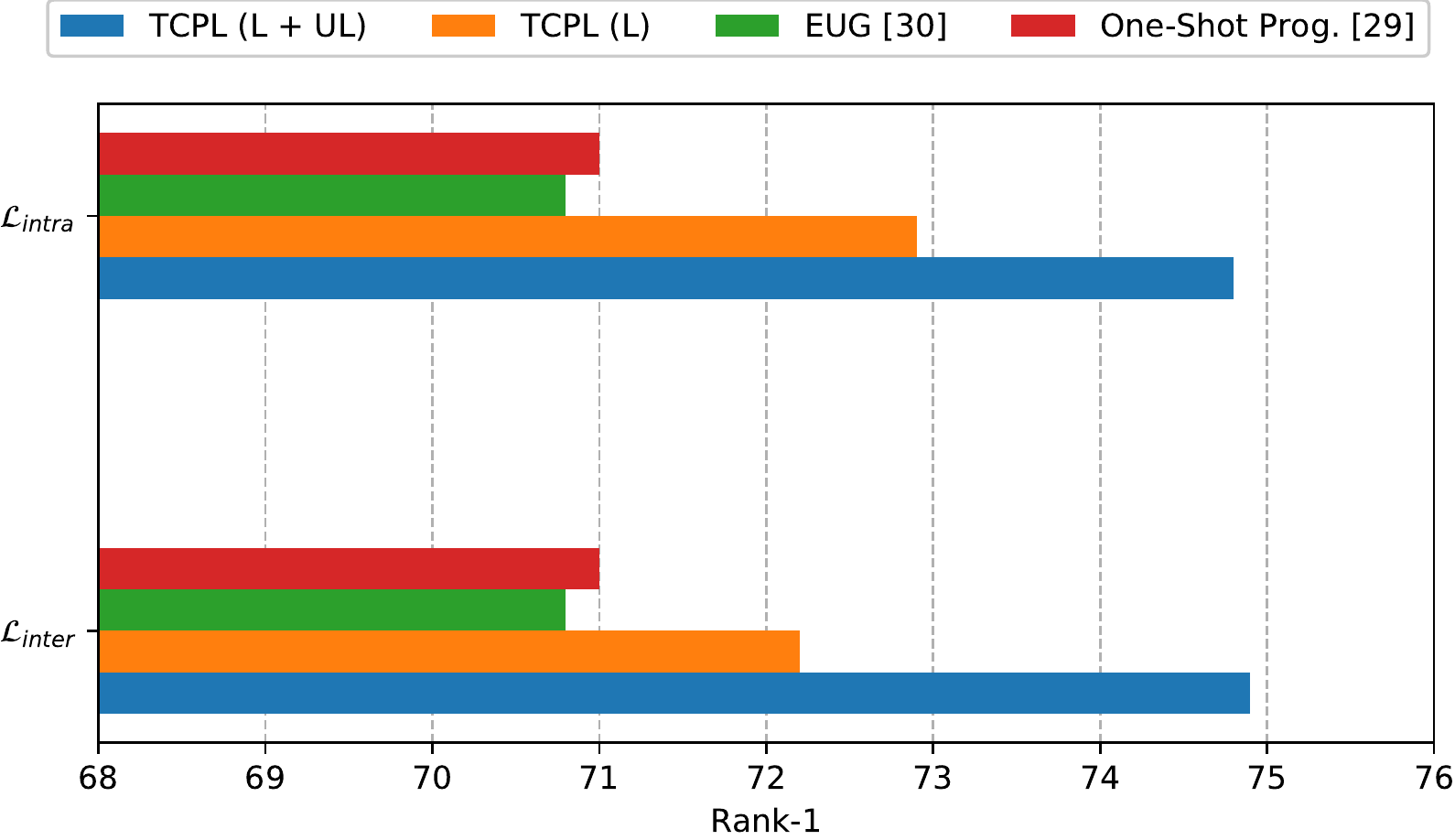} } 
\caption{{\bf Performance of TCPL by varying access to the unlabeled set.} (a) presents the Rank-1 acc. and (b) the mAP on DukeMTMC-VideoReID. Temporal consistency performs better than \cite{wu2018cvpr_oneshot,wu2019progressive} without using the entire unlabeled data, and improves even further when the unlabeled data is used. This demonstrates two things: (1) using the unlabeled data efficiently is important, (2) self-supervision can learn highly discriminative features. (L/UL denote the labeled/unlabeled set.)}
\label{fig:ablation} 
\end{figure}

\paragraph{{\bf Importance of maximally using the unlabeled data}.}
The ability to extract maximal information from the unlabeled data is at the core of TCPL. We demonstrate this in Fig. \ref{fig:ablation} by evaluating the losses on DukeMTMC-VideoReID with and without access to entire unlabeled data at each step of learning. 

The results confirm the two aspects of our hypothesis. Firstly, utilizing the entire unlabeled set at every step of learning improves performance. Secondly, self-supervision - even without access to the entire unlabeled set - learns better features and improves re-ID performance. TCPL, with access to only the labeled data, outperforms \cite{wu2019progressive} which accesses the entirety of the unlabeled set. This is a direct consequence of the ability of self-supervision to learn better features via consistency regularization, within and across camera views.

\begin{figure*}[t]
\centering
\subfloat[]{
	\label{subfig:hyper_r1}
	\includegraphics[width=0.45\textwidth]{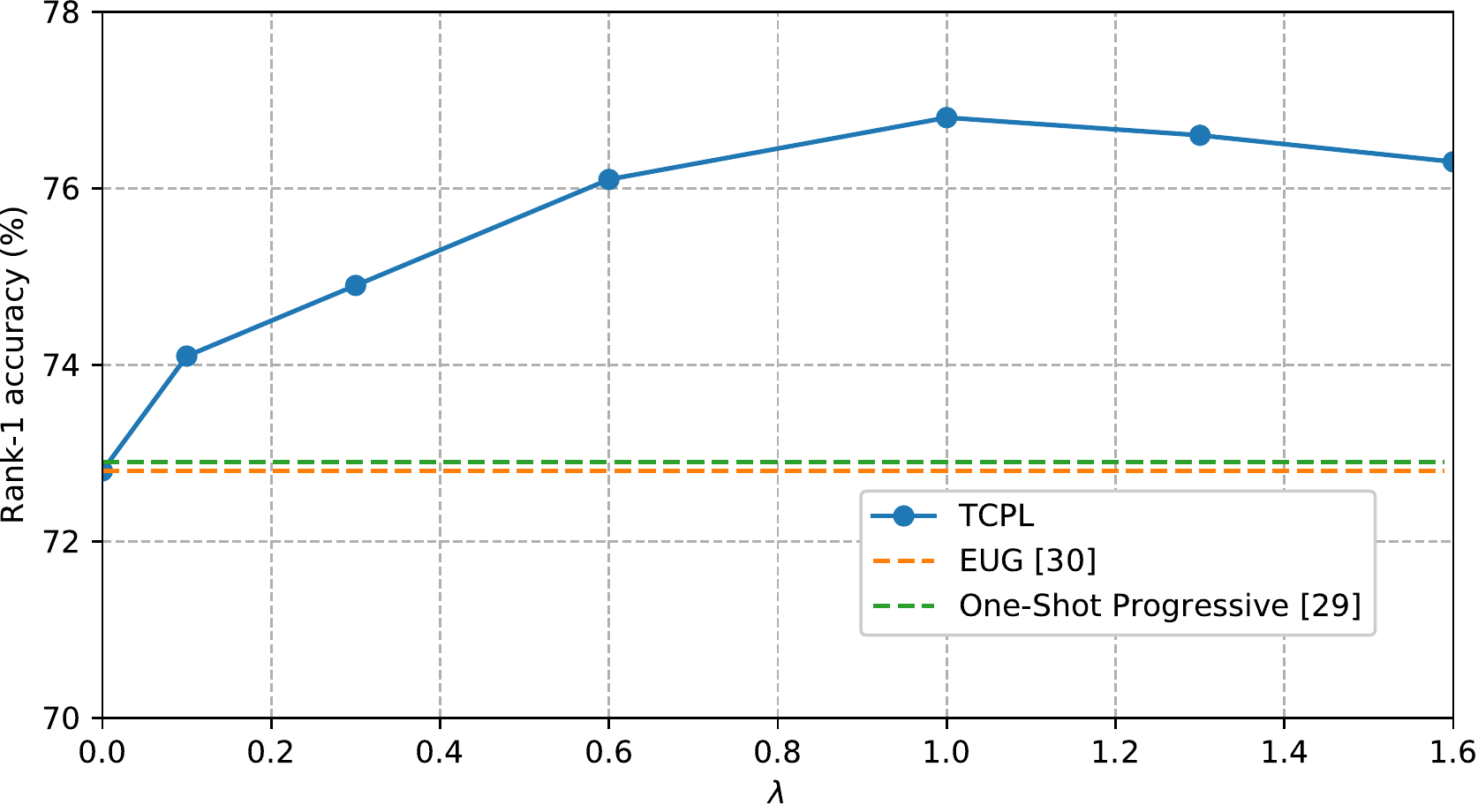} } 
\hfill
\subfloat[]{
	\label{subfig:hyper_map}
	\includegraphics[width=0.45\textwidth]{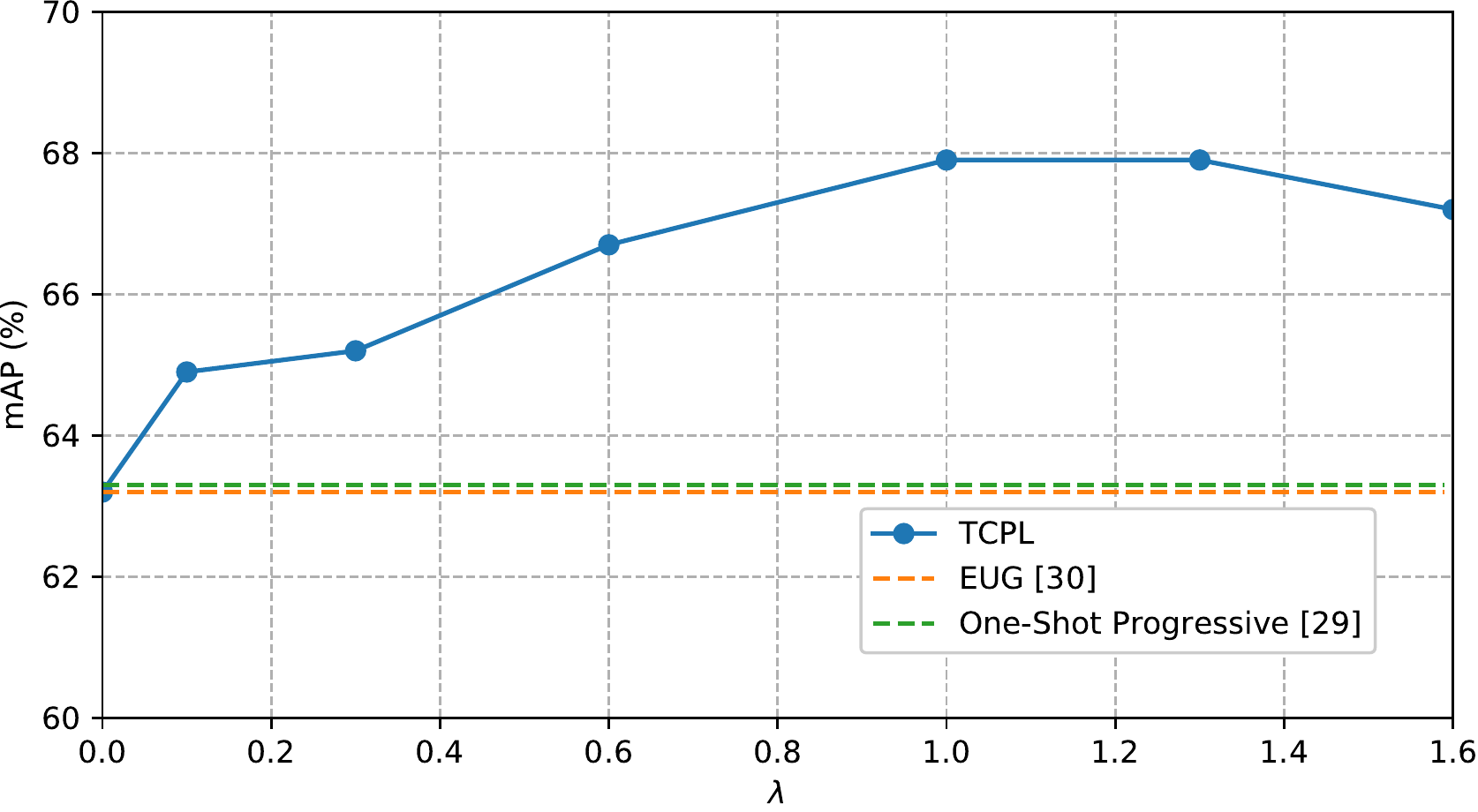} } 
\caption{{\bf Importance of temporal consistency.} (a) presents variations in Rank-1 accuracy on DukeMTMC-VideoReID by changing weights on temporal losses. Higher $\lambda$ represents more weight on the temporal losses. (b) presents the variations in mAP.}
\label{fig:hyper} 
\end{figure*}

\paragraph{{\bf Weight on the loss function}.} In our framework, we jointly optimize two types of losses - the cross-entropy loss and the temporal coherence losses ($\mathcal{L}_{\text{intra}},\mathcal{L}_{\text{inter}}$), as defined in Eqn. \ref{loss}, to learn the weights $\theta$ of the feature embedding $f_{\theta}(\cdot)$. We investigate the contributions of the temporal losses to the re-identification performance. In order to do that, we performed experiments with different values of $\lambda$ (higher value indicates larger weight on the temporal losses) and present the results on the DukeMTMC-VideoReID dataset in Fig. \ref{fig:hyper}. In general, increasing the weight improves performance, indicating the efficacy of self-supervision. As may be observed from the plot, the proposed method performs best with $\lambda=1$.

\begin{figure*}[t]
\centering
\subfloat[]{
	\label{subfig:mars_20}
	\includegraphics[width=0.45\textwidth]{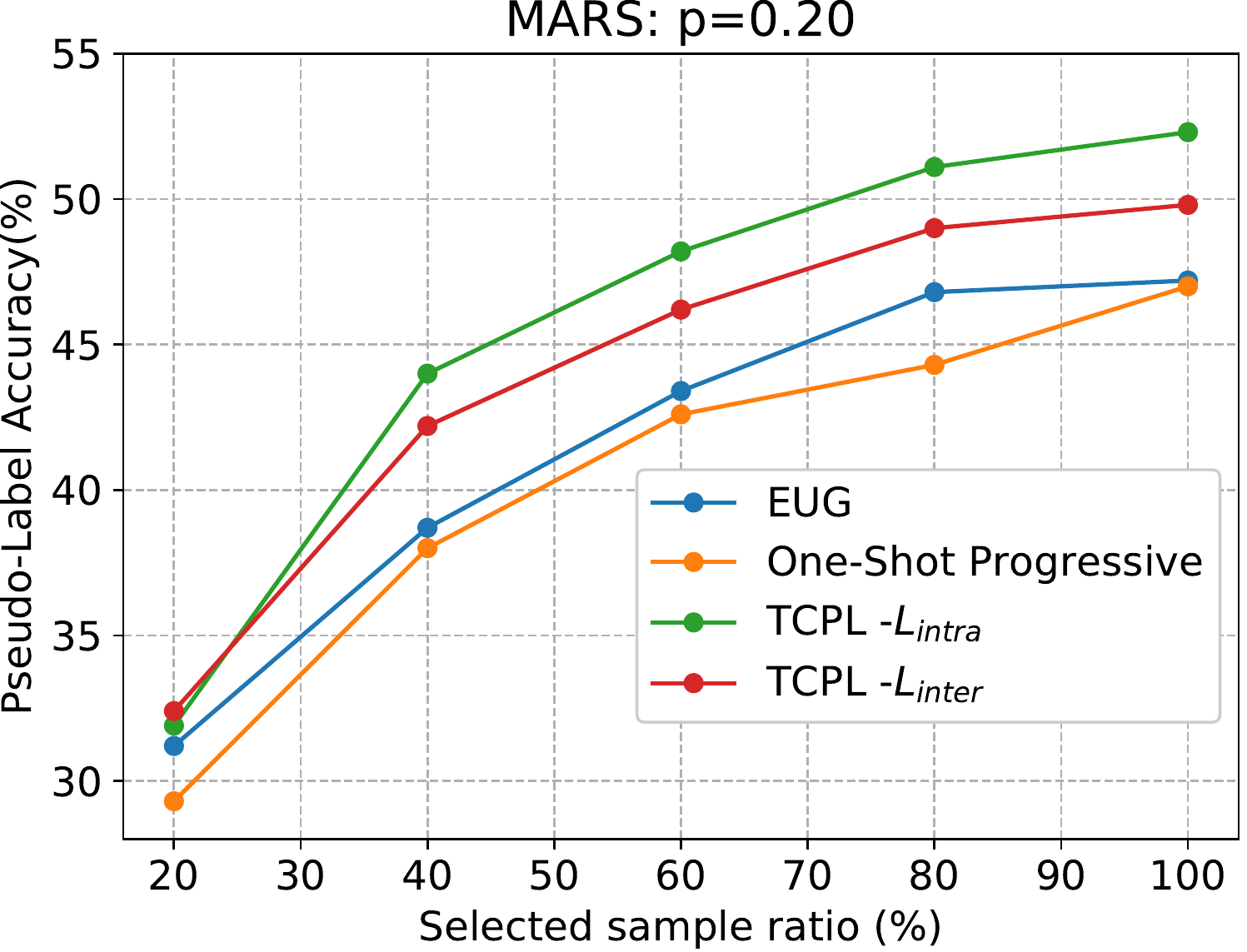} } 
\hfill
\subfloat[]{
	\label{subfig:duke_20}
	\includegraphics[width=0.45\textwidth]{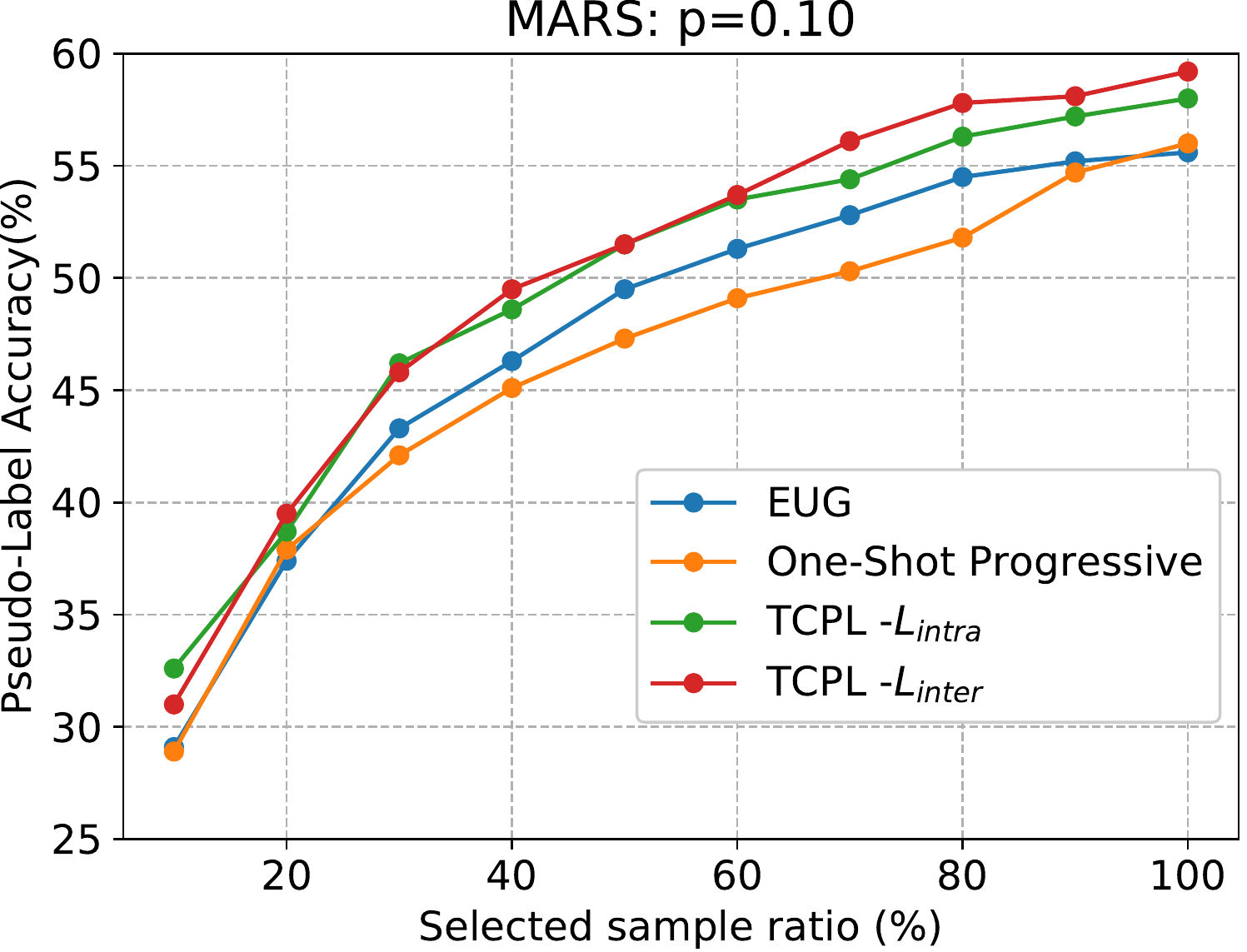} } 
\newline
\subfloat[]{
	\label{subfig:mars_10}
	\includegraphics[width=0.45\textwidth]{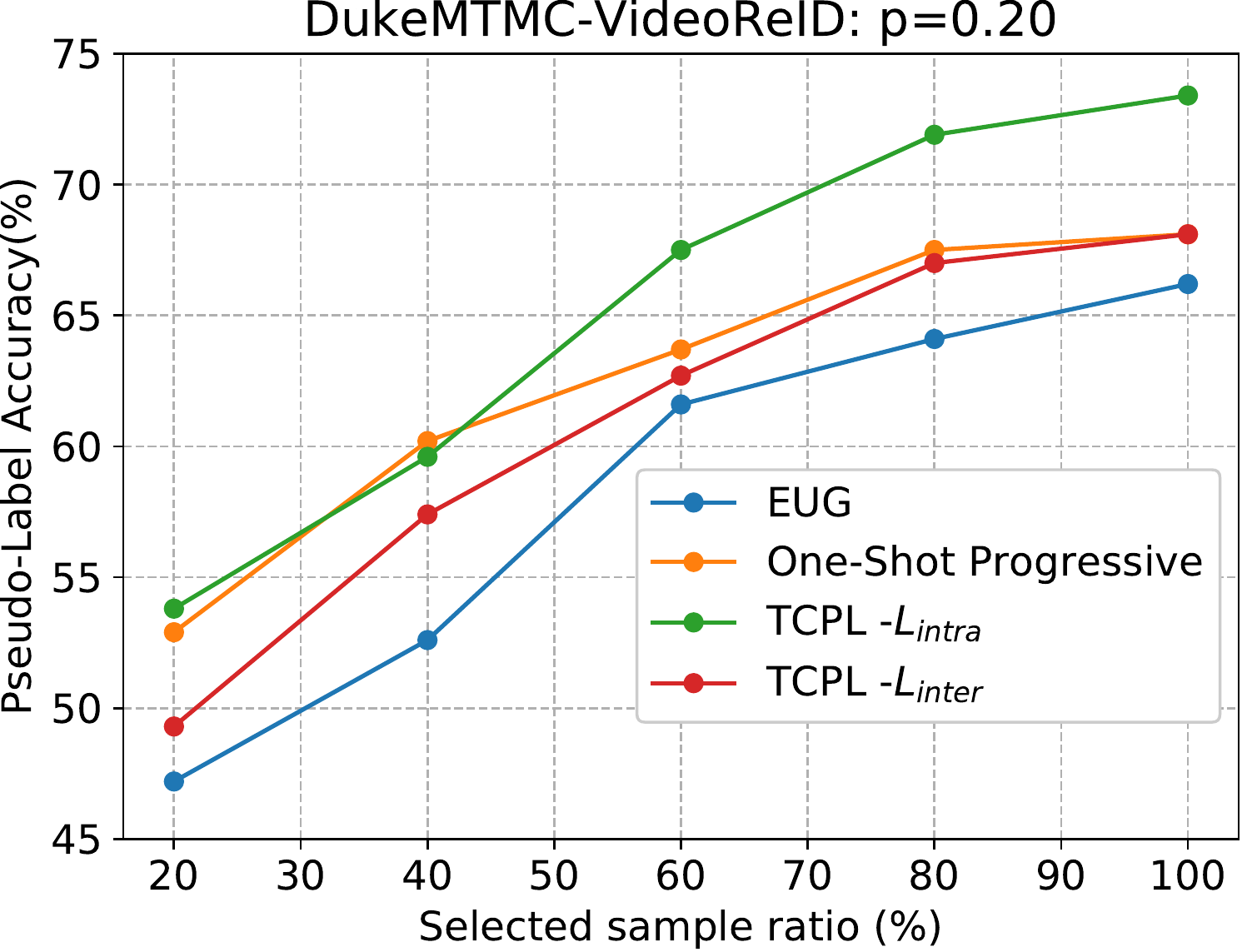} } 
\hfill
\subfloat[]{
	\label{subfig:duke_10}
	\includegraphics[width=0.45\textwidth]{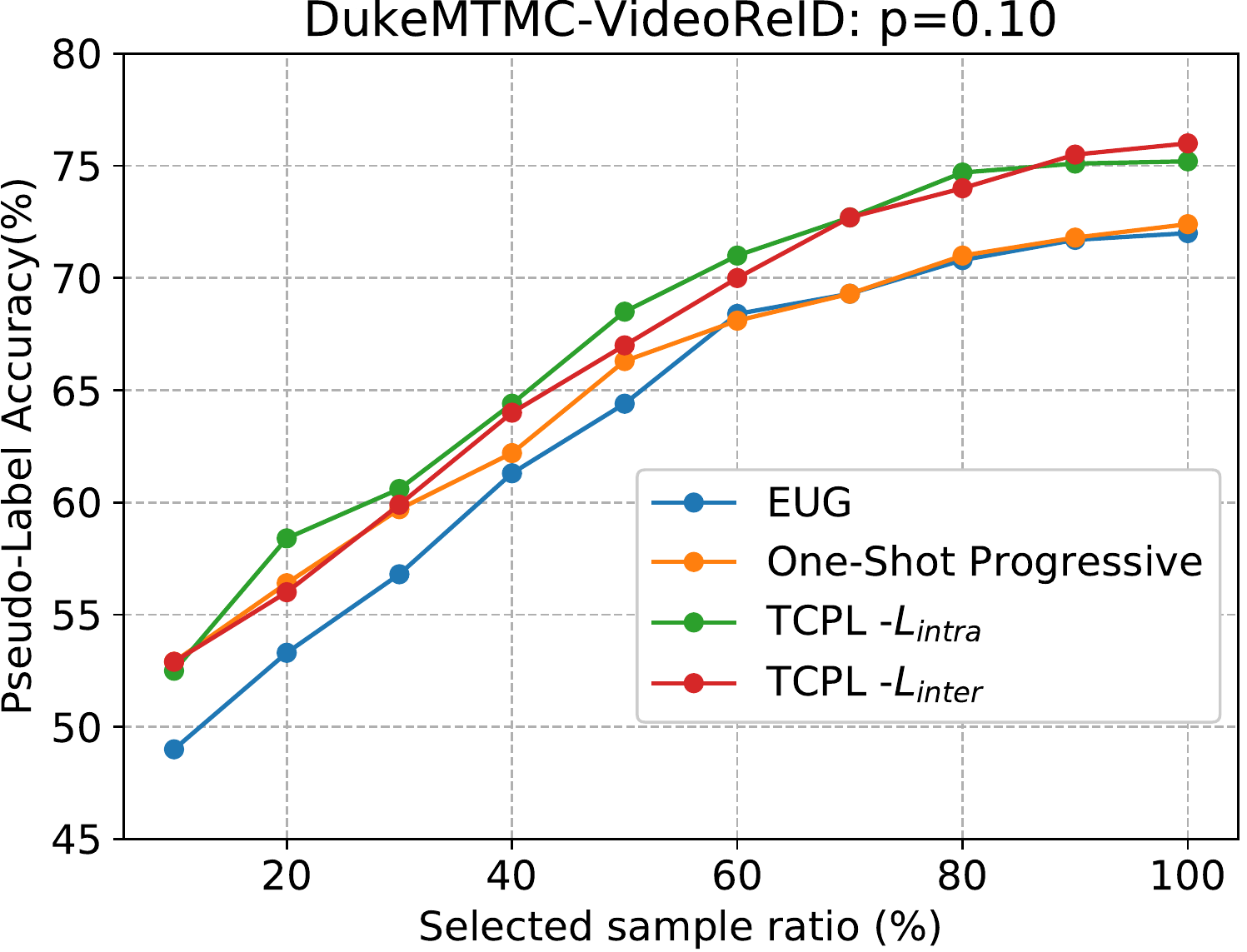} } 

\caption{{\bf Pseudo-label estimation.} Accuracy of pseudo-labels as enlarging factor $p$ is varied, on MARS [(a), (b)] and DukeMTMC-VideoReID [(c), (d)]}
\label{pl_acc}
\end{figure*}

\paragraph{{\bf Analysis over pseudo-label estimation}.} 
As a consequence of more discriminative feature learning using local consistency, TCPL is able to generate high quality labels for the unlabeled set. At $p=0.20$ and $p=0.10$, TCPL is able to achieve $\mathbf{8.2}\%$ and $\mathbf{4.0}\%$ improvement in label estimation respectively, on DukeMTMC-VideoReID, compared to EUG. On MARS, the improvement in estimation is $\mathbf{5.0}\%$ and $\mathbf{3.8}\%$ respectively. A visual representation of the improved pseudo-label estimation can be found in Fig. \ref{pl_acc}.

%%% CONCLUSION
\section{Conclusion}
In this paper, we introduce a new framework, Temporally Consistent Progressive Learning, which uses self-supervision via temporal coherence, in conjunction with one-shot labels, to learn a person re-ID model. Two novel temporal consistency losses, intra-sequence temporal consistency and inter-sequence temporal consistency, are at the core of this framework. These losses enable learning of richer and more discriminative representations. Our approach demonstrates the importance of using the unlabeled data efficiently and intelligently, an aspect of one-shot re-ID ignored by most previous works. Experiments on two challenging datasets establish our method as the state-of-the-art in the one-shot video person re-ID task. Future work will concentrate on extending the idea of temporal coherence to unsupervised person re-identification. 

\paragraph{\bf{Acknowledgments.}} We thank Sourya Roy, Sujoy Paul and Abhishek Aich for their assistance, advice and critique. The work was partially supported by NSF grant 1544969 and ONR grant N00014-19-1-2264.

\section*{Supplementary material}

\appendix 

\section{Evaluation on few-example setting}
Our method can be extended to the few example setting very easily, by acquiring more labeled samples before training. Thus, the labeled dataset may contain more than one tracklet per identity. We report the performance of our framework with varying ratios of labeled data in Table \ref{table:ssl}. 

\begin{table}
\caption{Comparison to the state-of-the-art supervised methods on MARS. We report performance in the semi-supervised (few-example) setting. The number in the bracket indicates the percentage of used labeled training data.}
\begin{center}
\setlength{\tabcolsep}{7pt}
\begin{tabular}{@{}llllll@{}}
\toprule[1.2pt]
Type                          & Method           & R-1  & R-5  & R-20                  & mAP  \\ \midrule[1.2pt]
\multirow{3}{*}{Supervised}      & ResNet50-3D \cite{gao2018revisiting}     & 82.9 & 93.7 & 96.8                  & 76.2 \\
                                 & IDTriplet \cite{hermans2017defense}       & 79.8 & 91.4 & \multicolumn{1}{c}{-} & 67.7 \\
                                 & Baseline (100\%) & 80.8 & 92.1 & 96.1                  & 67.4 \\ \midrule
\multirow{2}{*}{Semi-supervised} & Ours (10\%)      & 72.0 & 85.3 & 91.4                  & 56.5 \\
                                 & Ours (20\%)      & 78.2 & 89.9 & 94.4                  & 64.4 \\
\bottomrule[1.2pt]
\end{tabular}
\end{center}
\label{table:ssl}
\end{table}

On the MARS dataset, using only $20\%$ of the training data as the labeled set, our method achieves $78.2\%$ Rank-1 accuracy and $64.4\%$ mAP, which is very close to the fully supervised methods which utilize the entire training data with labels.
Although this setting requires more annotations than the one-shot task, it can easily achieve competitive results compared to the supervised methods. 

\section{Initial selection of tracklets}
We choose the labeled tracklets in a manner identical to the previous works \cite{wu2018cvpr_oneshot,liu2017stepwise}. More importantly, our method is designed to be robust to the selection of the labeled set - this is an advantage of our consistency losses, which promote discriminative feature learning regardless of labels. This robust behavior is demonstrated in Table \ref{tab:robust}
\setlength{\tabcolsep}{7pt}
\begin{table}[]
\caption{Results on Duke for $p=0.2$ across two random selections of the labeled set}
\centering
\begin{tabular}{@{}lll@{}}
\toprule
Split & R-1  & mAP  \\ \midrule
1     & 74.7 & 65.5 \\
2     & 74.4 & 65.2 \\ \bottomrule
\end{tabular}
\label{tab:robust}
\end{table}

\section{Analysis on range parameter $r$}
The range parameter $r$ plays an important role in the context of the inter-sequence consistency criterion. Choosing $r$ too high will lead to sampling of easy negatives, which will not contribute too much to learning (zero gradients). On the flip side, choosing $r$ too low can lead to positives being interpreted as negatives, which can hamper learning. We demonstrate this behavior in Table \ref{tab:range}.

\setlength{\tabcolsep}{7pt}
\begin{table}
\caption{Performance on MARS for $\mathcal{L}_{\mathrm{inter}}$ and $p=0.20$ as the range parameter $r$ is varied.}
\centering
\begin{tabular}{@{}lll@{}}
\toprule
$r$ & R-1 & mAP \\ \midrule
1 & 53.1 & 30.2 \\
2 & 53.6 & 30.6 \\
3 & 50.8 & 27.9 \\ \bottomrule
\end{tabular}
\label{tab:range}
\end{table}

\section{Additional results}
\begin{figure}[h]
\centering

\subfloat[]{
	\label{subfig:duke1}
	\includegraphics[width=0.23\textwidth]{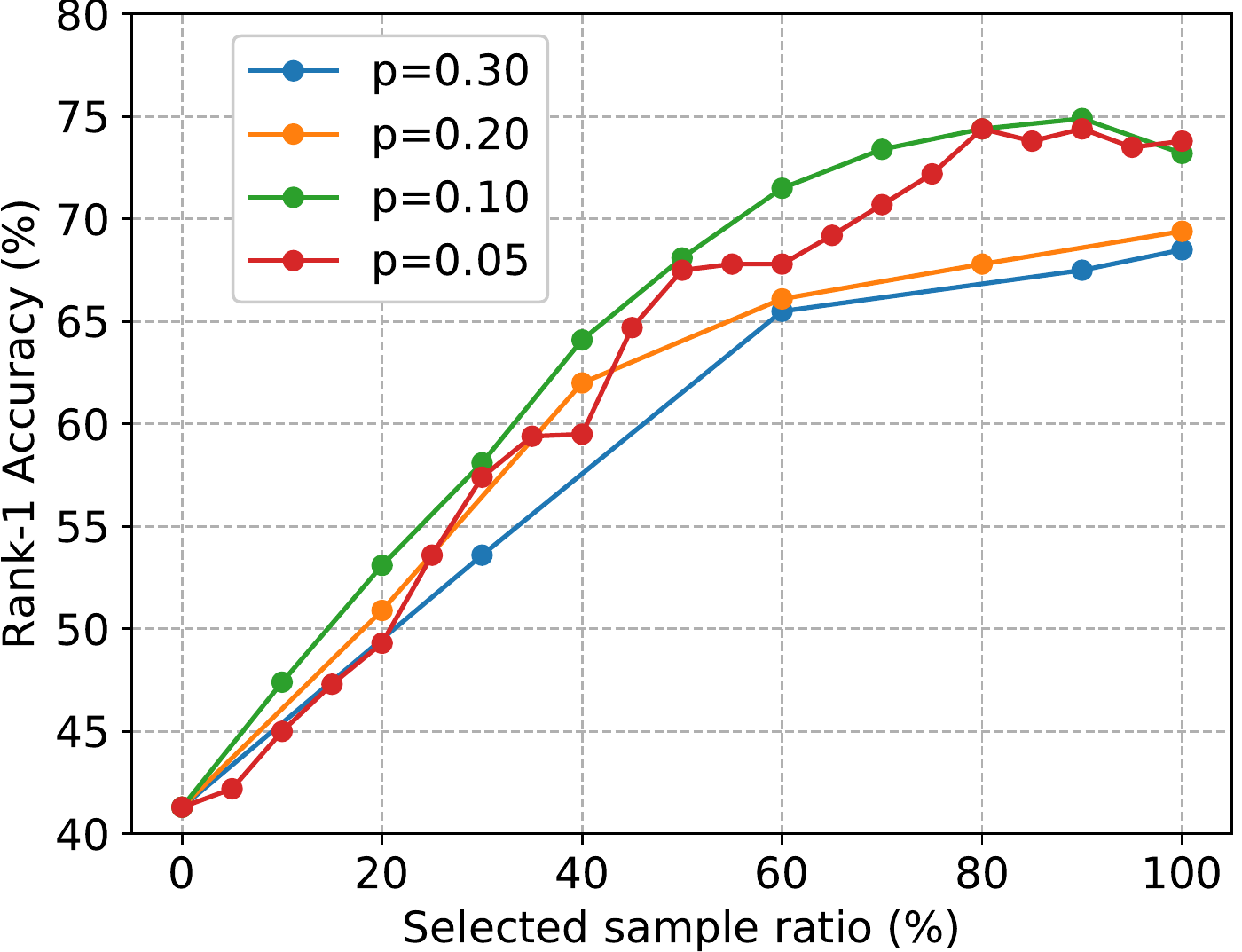} } 
\hfill
\subfloat[]{
	\label{subfig:duke2}
	\includegraphics[width=0.23\textwidth]{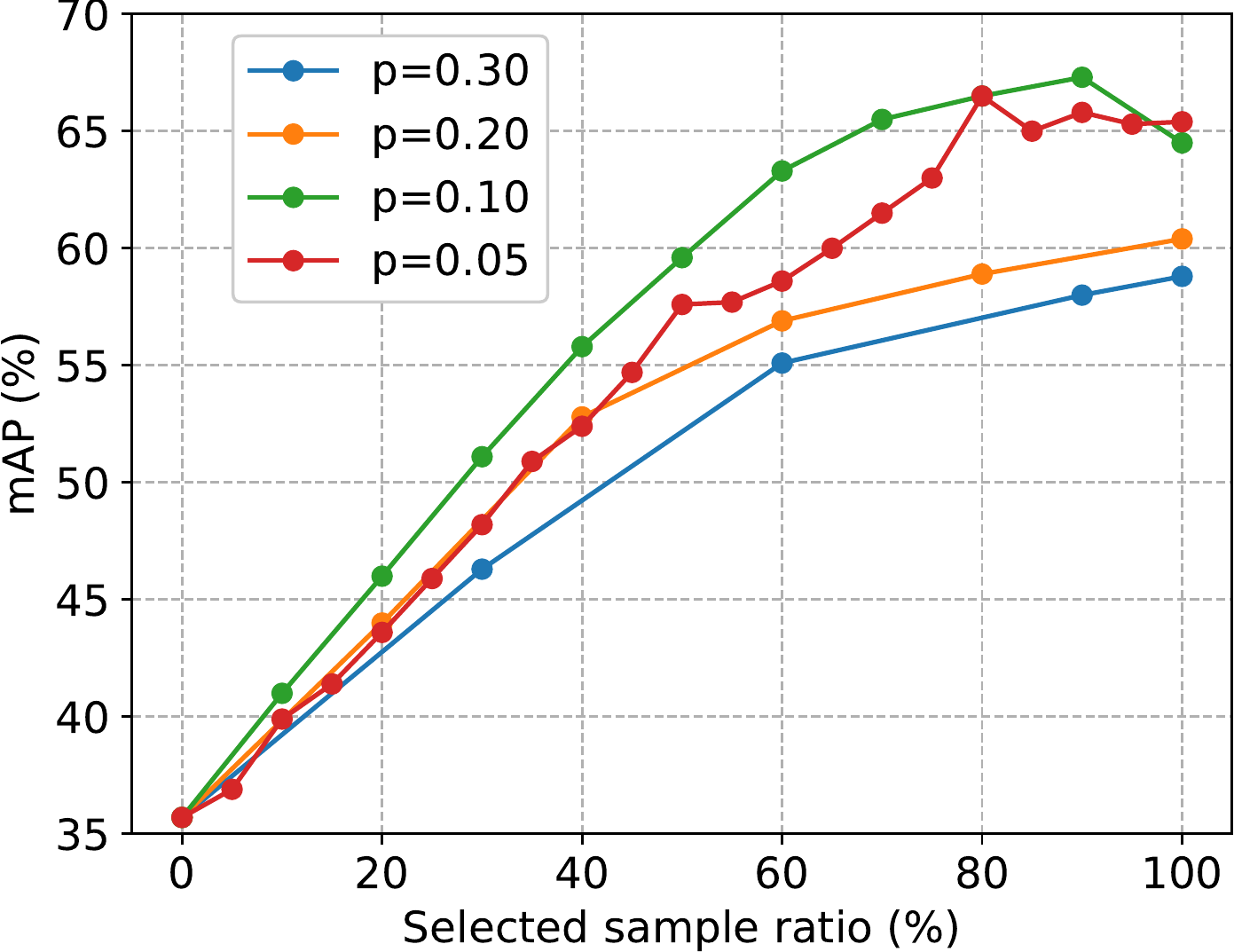} } 
\hfill
\subfloat[]{
	\label{subfig:duke3}
	\includegraphics[width=0.23\textwidth]{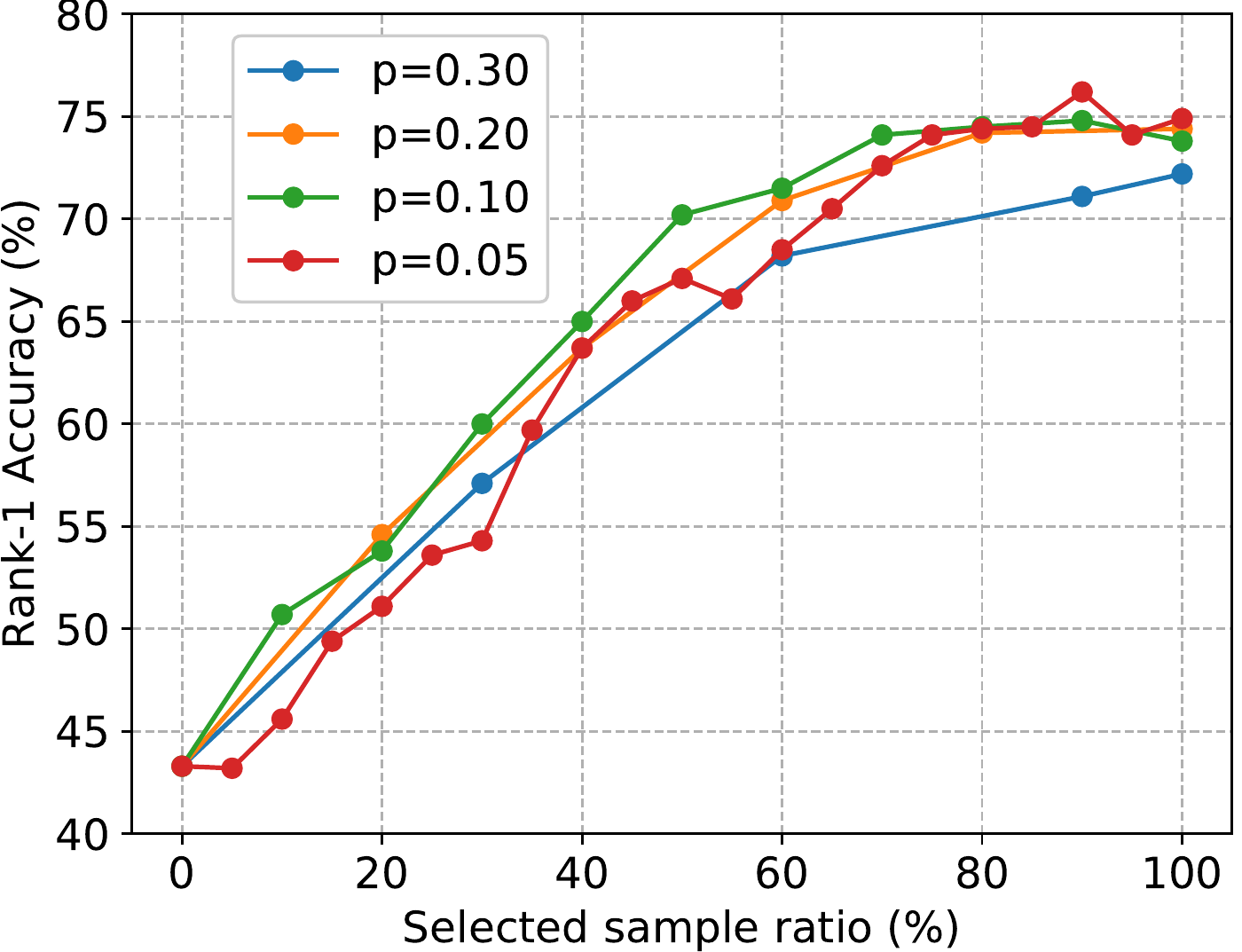} } 
\hfill
\subfloat[]{
	\label{subfig:duke4}
	\includegraphics[width=0.23\textwidth]{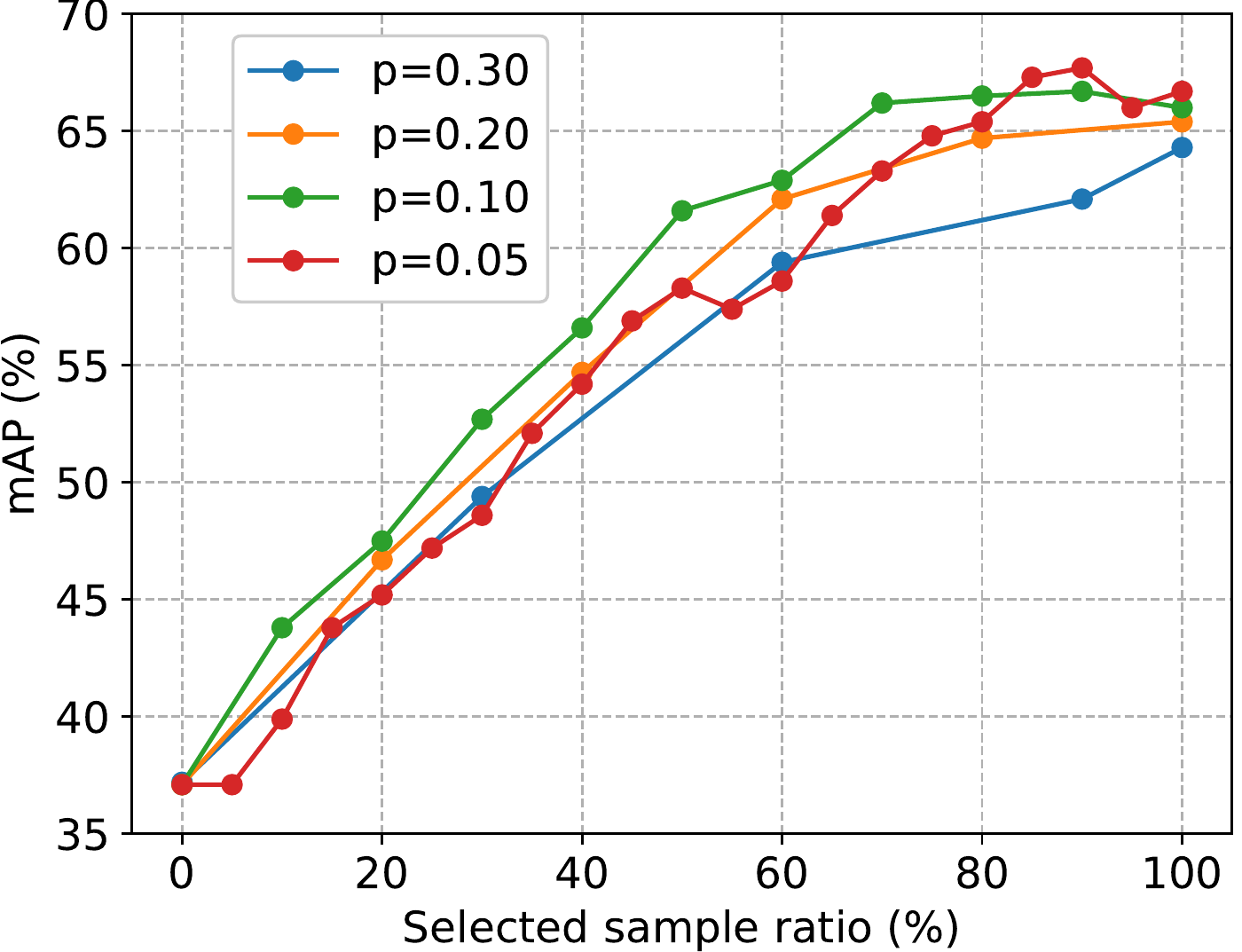} } 
	
\caption{Comparison with different values of enlarging factor on DukeMTMC-VideoReID. Figures (a) and (b) represent the Rank-1 accuracy and mAP while using $\mathcal{L}_{\text{inter}}$. Figures (c) and (d) represent the Rank-1 accuracy and mAP while using $\mathcal{L}_{\text{intra}}$.}
\label{fig:duke_learning_curves}
\end{figure}

In Table \ref{duke_results}, we present the case when the enlarging factor $p=0.30$. This indicates a very aggressive incorporation of pseudo-labels and increases the chance for erroneous label estimation. However, even in this case TCPL is able to perform better than competing methods. Especially on the DukeMTMC-VideoReID dataset, TCPL outperforms \cite{wu2019progressive} in mAP by $\mathbf{8}\%$. In Figure \ref{fig:duke_learning_curves}, we present the learning curves on DukeMTMC-VideoReID.
\setlength{\tabcolsep}{7pt}
\begin{table*}[t]
\caption{One-Shot Performance for the enlarging parameter $p=0.30$. The best and second best results are in \textcolor{red}{red}/\textcolor{blue}{blue} respectively.}
\begin{center}
\begin{tabular}{lllllll@{}}
\toprule[1.2pt]
Dataset        &Setting                  & Methods               & R-1  & R-5  & R-20 & mAP  \\ 
\midrule[1.2pt]
\addlinespace
\multirow{4}{*}{DukeMTMC} & \multirow{4}{*}{$p = 0.30$} & EUG  \cite{wu2018cvpr_oneshot}                 & 63.8 & 78.6 & 87.0 & 54.6 \\
                         & & One-Shot Progressive \cite{wu2019progressive} & 66.1 & 79.8 & 88.3 & 56.3 \\
                         & & TCPL -$\mathcal{L}_{\text{intra}}$           & \textcolor{red}{72.2} & \textcolor{red}{83.2} & \textcolor{red}{90.3} & \textcolor{red}{64.3} \\
                         & & TCPL -$\mathcal{L}_{\text{inter}}$            & \textcolor{blue}{68.5} & \textcolor{blue}{80.8} & \textcolor{blue}{88.6} & \textcolor{blue}{58.8} \\ 
                         \midrule
\multirow{4}{*}{MARS} &\multirow{4}{*}{$p = 0.30$} & EUG \cite{wu2018cvpr_oneshot}              & 42.8 & 56.5 & 67.2 & 21.1 \\
                         & & One-Shot Progressive \cite{wu2019progressive} & 44.5 & 58.7 & \textcolor{red}{70.6} & 22.1 \\
                         & & TCPL -$\mathcal{L}_{\text{intra}}$            & \textcolor{blue}{45.3} & \textcolor{blue}{57.6} & 
                         66.7 & 
                         \textcolor{blue}{23.8} \\
                         & & TCPL -$\mathcal{L}_{\text{inter}}$            & \textcolor{red}{45.7} & \textcolor{red}{59.6} & \textcolor{blue}{69.3} & \textcolor{red}{23.9}\\                       
 \bottomrule[1.2pt]
\end{tabular}
\end{center}

\label{duke_results}
\end{table*}

\section{Dataset overview}
The {\bf MARS} dataset \cite{zheng2016mars} is the largest video person re-identification dataset for the person and was collected in a university campus. The dataset contains $17503$ tracklets for $1261$ identities and $3248$ distractor tracklets, which are captured by six cameras. The dataset is split into $625$ identities for training and $636$ identities for testing. Every identity in the training set has approximately $13$ video tracklets on average and $800$ frames on average. The bounding boxes are detected and tracked using the Deformable Part Model (DPM) and GMMCP tracker. 

The {\bf DukeMTMC} dataset \cite{ristani2016performance} was released with the aim of developing multi-camera tracking algorithms. The dataset was captured in outdoor scenes with noisy background and suffers from illumination, pose, and viewpoint change and occlusions. The {\bf DukeMTMC-VideoReID} \cite{wu2018cvpr_oneshot} is a subset of the DukeMTMC dataset created for video re-identification. The dataset is manually annotated and each identity has a singular tracklet under a camera. The dataset contains $702$ identities for training, $702$ identities for testing, and $408$ identities as the
distractors. In total there are $369,656$ frames of $2,196$ tracklets for training, and $445,764$ frames of $2,636$ tracklets for testing and distractors.
\begin{figure}[h]
\centering
\subfloat[DukeMTMC-VideoReID]{
	\label{subfig:duke}
	\includegraphics[width=0.46\textwidth]{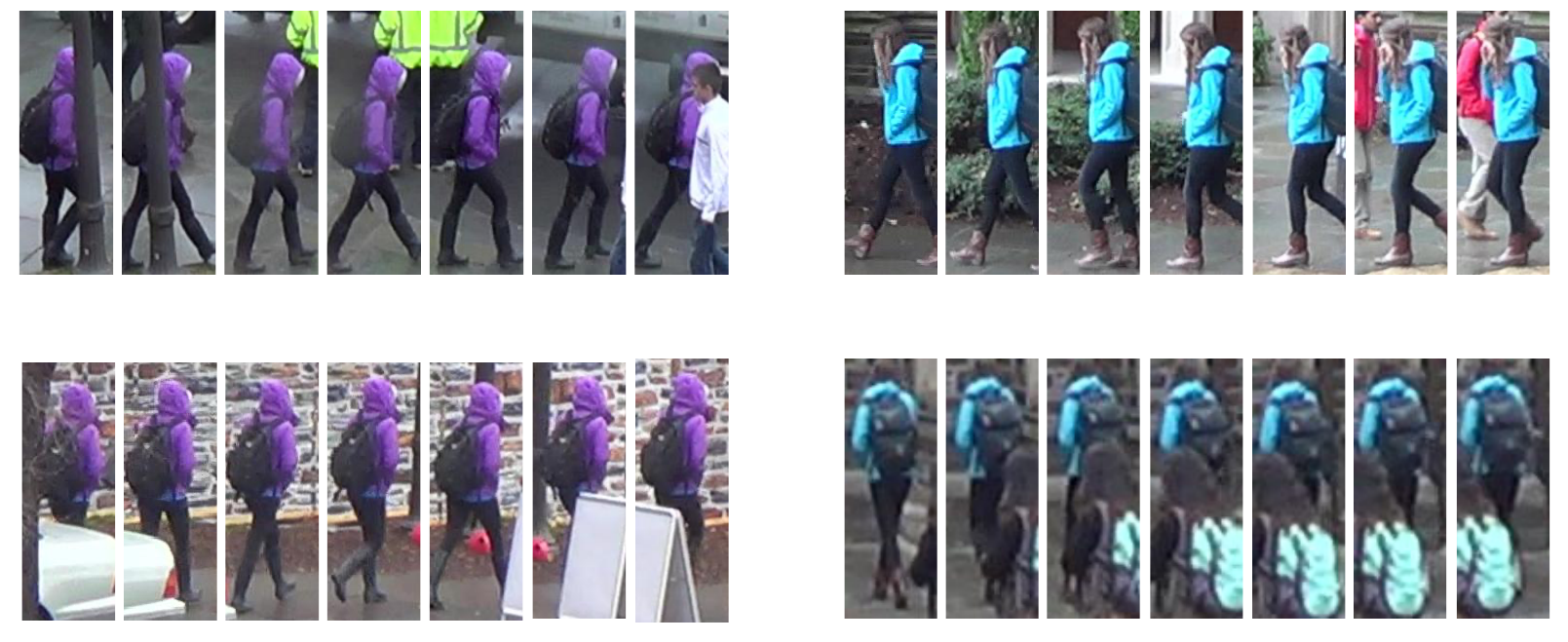}} 
\hfill
\subfloat[MARS]{
	\label{subfig:mars}
	\includegraphics[width=0.46\textwidth]{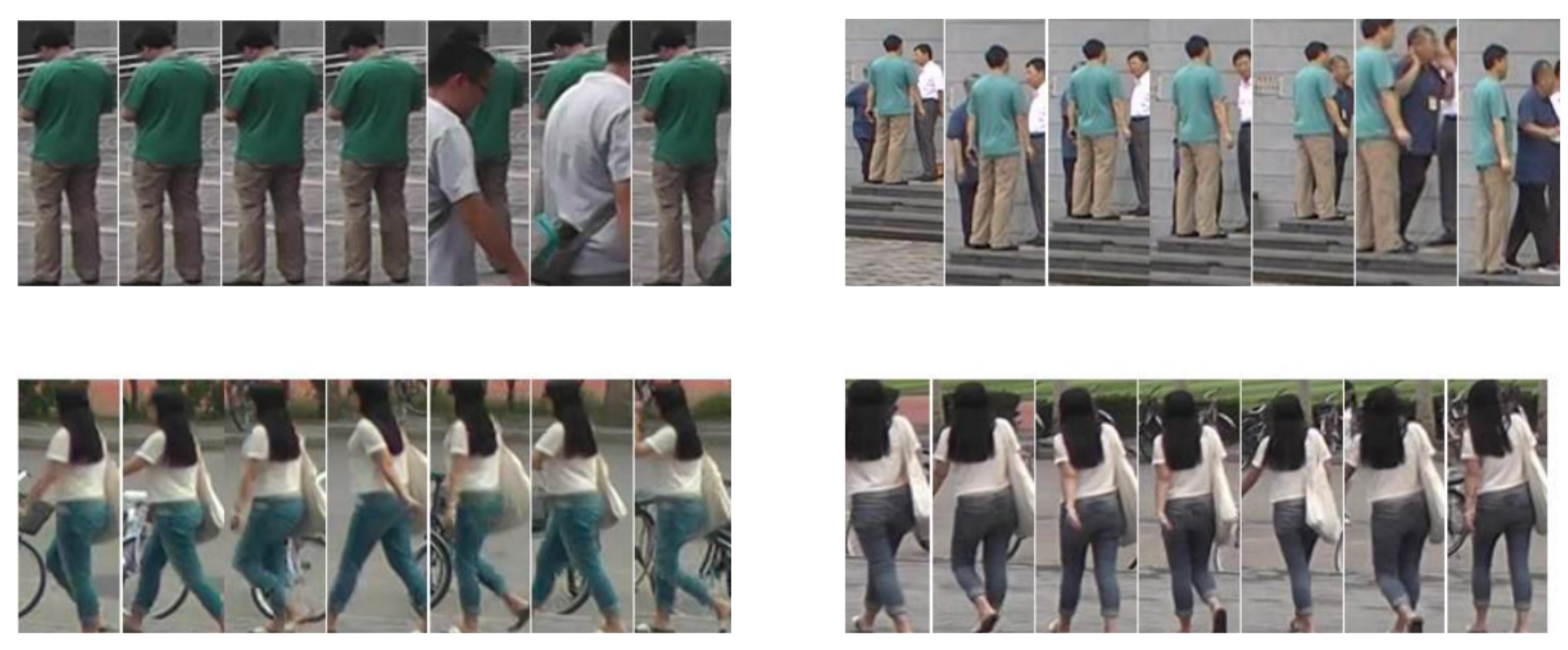}} 

\caption{A total of 8 sample tracklets from the two datasets used in our experiments. Each column represents a distinct individual, with the rows denoting two different views of the same person from two different cameras. We can see that across cameras, the tracklets of the same person vary significantly due to changes in illumination, occlusion etc. Even within a tracklet, the background varies significantly.}
\label{fig:datasets}
\end{figure}

\section{Implementation details}
We use PyTorch \cite{paszke2019automatic} for all experiments. For our model, we use a ResNet-50 \cite{he2016deep} pre-trained on ImageNet \cite{deng2009imagenet} - the last classification layer removed and a fully-connected layer with batch normalization \cite{ioffe2015batch} and a classification layer are added at the end of the model. We adopt stochastic gradient descent (SGD)  with momentum $0.5$ and weight decay $0.0005$ to optimize the parameters for $70$ epochs, with batch size $16$ in each iteration. We set $\lambda=1$ in for the DukeMTMC-VideoReID dataset and $\lambda=0.8$ for the MARS dataset (due to the huge disparity in the number of labeled and unlabeled tracklets as a result of fragmentation in MARS). The learning rate is initialized to $0.1$. In the last 15 epochs, to stabilize the model training and prevent overfitting, we change the learning rate to $0.01$ and set $\lambda = 0$.

\bibliographystyle{splncs04}
\bibliography{main}
\end{document}